\documentclass[10pt,twocolumn,letterpaper]{article}

\usepackage{iccv}
\usepackage{times}
\usepackage{epsfig}
\usepackage{graphicx}
\usepackage{amsmath}
\usepackage{amssymb}
\usepackage{enumitem}

\usepackage{multirow}
\usepackage{booktabs}
\usepackage{amsthm}
\usepackage{multirow}
\usepackage{algorithm} 
\usepackage{algpseudocode}  
\newtheorem{hypothesis}{Hypothesis}
\newtheorem{proposition}{Proposition}
\usepackage[accsupp]{axessibility}  

\usepackage[pagebackref=true,breaklinks=true,letterpaper=true,colorlinks,bookmarks=false]{hyperref}

\iccvfinalcopy 

\def\iccvPaperID{11371} 

\ificcvfinal\fi

\begin{document}

\title{Partition Speeds Up Learning Implicit Neural Representations Based on Exponential-Increase Hypothesis}



\author{
  Ke Liu$^1$, Feng Liu$^2$, Haishuai Wang$^{1}$\thanks{Corresponding author: Haishuai Wang (haishuai.wang@zju.edu.cn)}~, Ning Ma$^1$, Jiajun Bu$^1$, Bo Han$^3$  \vspace{0.2cm} \\ 
  $^1$ Zhejiang Provincial Key Laboratory of Service Robot, \\ College of Computer Science, Zhejiang University, Hangzhou, China \\
  $^2$School of Computing and Information Systems, The University of Melbourne, Australia \\
  $^3$Department of Computer Science, Hong Kong Baptist University, Hong Kong SAR, China\vspace{0.1cm} \\
  \texttt{\{keliu99, haishuai.wang, ma\_ning, bjj\}@zju.edu.cn,} \\
  \texttt{fengliu.ml@gmail.com, bhanml@comp.hkbu.edu.hk}
}

\maketitle
\ificcvfinal\thispagestyle{empty}\fi

\begin{abstract}
\textit{Implicit neural representations} (INRs) aim to learn a \textit{continuous function} (i.e., a neural network) to represent an image, where the input and output of the function are pixel coordinates and RGB/Gray values, respectively. 
However, images tend to consist of many objects whose colors are not perfectly consistent, resulting in the challenge that image is actually a \textit{discontinuous piecewise function} and cannot be well estimated by a continuous function. 
In this paper, we empirically investigate that if a neural network is enforced to fit a discontinuous piecewise function to reach a fixed small error, the time costs will increase exponentially with respect to the boundaries in the spatial domain of the target signal.
We name this phenomenon the \textit{exponential-increase} hypothesis. Under the \textit{exponential-increase} hypothesis, learning INRs for images with many objects will converge very slowly.
To address this issue, we first prove that partitioning a complex signal into several sub-regions and utilizing piecewise INRs to fit that signal can significantly speed up the convergence.
Based on this fact, we introduce a simple partition mechanism to boost the performance of two INR methods for image reconstruction: one for learning INRs, and the other for learning-to-learn INRs. 
In both cases, we partition an image into different sub-regions and dedicate smaller networks for each part. In addition, we further propose two partition rules based on regular grids and semantic segmentation maps, respectively.
Extensive experiments validate the effectiveness of the proposed partitioning methods in terms of learning INR for a single image (ordinary learning framework) and the learning-to-learn framework. Code is released \href{https://github.com/1999kevin/INR-Partition.git}{here}.

\end{abstract}

\vspace{-0.5cm}
\section{Introduction}
\label{section:intro}

Recently, an innovative model for data/signal representation called implicit neural representations (INRs) has aroused researchers' great attention, due to their remarkable visual performance in computer vision tasks, including image generation~\cite{sitzmann2020implicit,martel2021acorn,skorokhodov2021adversarial,dupont2022data} and novel views synthesis~\cite{mildenhall2021nerf}. To fit such an \emph{implicit neural representation} for a 2D image, we usually learn a continuous function formalized by a neural network, which takes space coordinates $x \in \mathbb{R}^2$ as input and outputs the color values at the queried coordinate ($y \in \mathbb{R}^3$ if RGB and $y \in \mathbb{R}$ if gray).

\begin{figure*}[!t]
\centering
\includegraphics[scale=0.57]{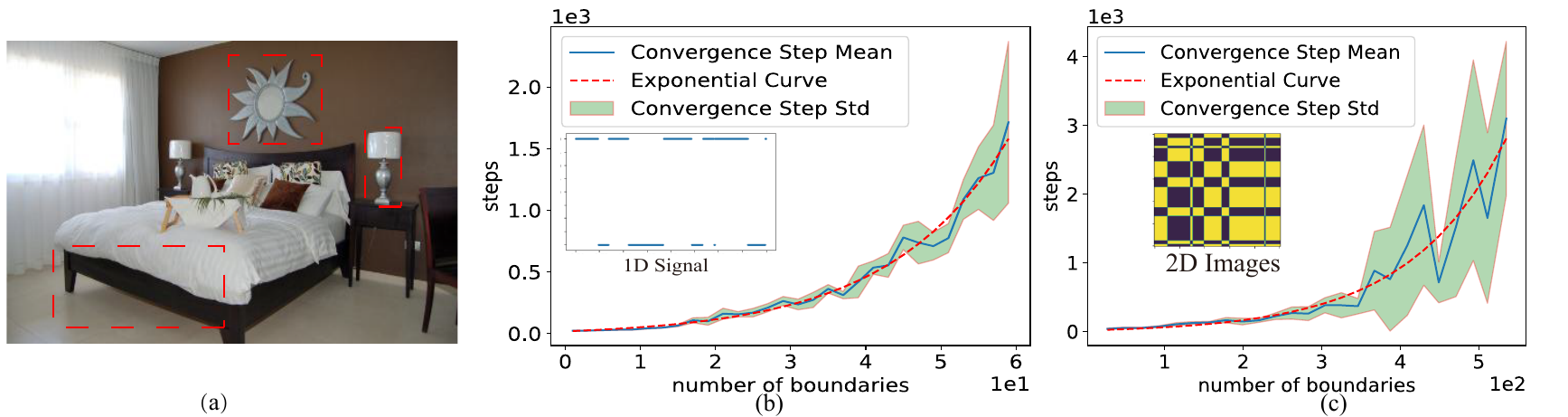}
\caption{(a) Discontinuous parts exist obviously in regions with red boxes, which motivates us to use piecewise functions to represent the complex signals. (b) \& (c) We dedicate a single neural network to fit the 1D and 2D synthetic signals with $N$ boundaries. The results show that the relation between the convergence step $n$ and the number of boundaries $N$ align with the exponential function $n \propto O(p^{N})$, where $p=1.0656$ for 1D synthetic signals and $p=1.00815$ for 2D synthetic signals. The detailed experiment is presented in Appendix~\ref{section:motivation_experiment}.}
\label{fig:motivation}
\vspace{-1em}
\end{figure*}

However, in-the-wild images are actually \emph{discontinuous piecewise functions}. They consist of discrete objects with not perfectly consistent colors (as shown in Figure~\ref{fig:motivation}(a)). Large gradients exist on the boundaries between two discontinuous parts, preventing the neural network from converging to a small error when fitting images. 
To study the above issue, related research called ``spectral bias''~\cite{rahaman2019spectral,tancik2020fourier} has proved that neural networks prioritize learning the low-frequency components. Yet, they only describe this phenomenon from the view of the implicit frequency domain and do not propose a quantitative relation between the convergence rate and the attribute of the target signal.

In this paper, we first re-examine the above phenomenon from the explicit spatial domain and empirically investigate a quantitative relation: the time complexity of fitting a discontinuous piecewise function with a neural network would increase exponentially with respect to the number of boundaries. For example, in Figure~\ref{fig:motivation}(b) and \ref{fig:motivation}(c), we use SIREN MLPs~\cite{sitzmann2020implicit} to fit 1D synthetic signals and 2D synthetic signals where $N$ boundaries exist in their spatial domain. We then explore the relation between the required convergence step $n$ and the number of boundaries $N$, and find that the relation curves align with the exponential function.
We call this phenomenon the \emph{exponential-increase} hypothesis. 
Under this hypothesis, the optimization process of fitting a high-resolution in-the-wild image with a single continuous INR will converge at a \emph{slow rate}.

Based on the exponential-increase hypothesis, we mathematically prove that partitioning images into several parts and learning INRs within each part can reduce the exponential complexity to linear complexity and significantly decrease the convergence time. In light of this fact, we propose partition-based INR methods and utilize partition in two INR frameworks: one for learning INRs, and the other for learning-to-learn INRs. Specifically, in both frameworks, we partition an image into different sub-regions based on particular rules and dedicate smaller networks for each sub-region. We also propose two partition rules: one is based on regular grids, and the other is based on semantic segmentation maps. Both of them can speed up the convergence of learning INRs as well as learning-to-learn INRs. In summary, the contributions of this work are as follows.

\begin{itemize}[itemsep=2pt,topsep=0pt,parsep=0pt]
    \item From the view of spatial domains, we investigate the exponential relation between the network convergence rate and the number of boundaries in the target signals, namely the exponential-increase hypothesis. 
    \item Based on the exponential-increase hypothesis, we mathematically prove that partition reduces the exponential complexity of fitting all boundaries to the linear complexity of fitting separate regions.  
    \item We propose partition-based learning and learning-to-learn INRs frameworks for image reconstruction task. We also propose two partition rules that are based on regular grids or semantic segmentation maps.  
    \item Extensive experiments on image reconstruction show that (i) partition boosts learning INRs framework to faster convergence, (ii) partition boosts learning-to-learn INRs framework to better reconstruction performance with fixed optimization steps.
    
\end{itemize}

\section{Related Work}

\textbf{Implicit Neural Representations.}
Implicit Neural Representations (INRs)~\cite{xie2022neural} are emerging topics of interest in the artificial intelligence community. By mapping a coordinate $x$ to a quantity with a neural network (e.g., MLP), these continuous representations have shown great potential in 
3D scene reconstruction~\cite{hao2020dualsdf,lou2019realistic,park2019deepsdf,michalkiewicz2019implicit},
digital humans tasks~\cite{yenamandra2021i3dmm,saito2019pifu,saito2021scanimate},
2D images generation~\cite{qin2022hilbert,sitzmann2020implicit,skorokhodov2021adversarial,zhou2021distilling},
3D shape and appearance generation~\cite{lou2021real,schwarz2020graf,niemeyer2021giraffe,mildenhall2021nerf},
video representation~\cite{chen2021nerv},
physics-informed problems~\cite{raissi2019physics,pfrommer2020contactnets} and so on.
A lot of works have been conducted on different aspects of INRs, such as the prior learning and conditioning~\cite{sitzmann2019scene, tretschk2020patchnets,wang2022neural}, the computation and memory efficiency~\cite{meister2021survey}, the expression capacity~\cite{rebain2021derf,yuce2022structured}, the edit ability ~\cite{sitzmann2019scene} and the generalization across different samples~\cite{sitzmann2020metasdf,tancik2021learned}.

\textbf{Partition Techniques.} When scaling up to signals with large domains, INRs always fail due to the high non-linearity of mapping function~\cite{ren2013global} and heavy time consumption. Thus, partition are extensively employed, e.g. Voronoi spatial decomposition by DeRF~\cite{rebain2021derf}, distillation for training thousands of MLPs by KiloNeRF~\cite{reiser2021kilonerf}, multiscale block-coordinate decomposition by ACRON~\cite{martel2021acorn}, scalable large-scale NeRF~\cite{tancik2022block,turki2022mega}.
Although these works have a good effect on representing large-scale images or scenes, they seldom discuss why partition improves the training efficiency of learning single INR and do not discuss the effect of partition on the learning-to-learn INRs framework.

\begin{figure*}[!t]
\centering
\includegraphics[scale=0.75]{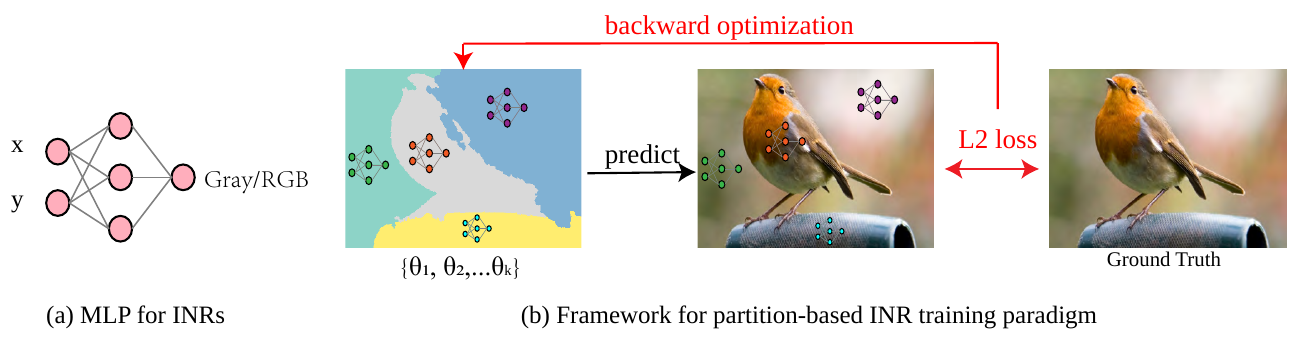}
\caption{(a) MLP architecture for image INR. The inputs are the coordinates $(x,y) \in \mathbb{R}^2$ and the outputs are Gray/ RGB values. (b) Framework for partition-based Learning INRs. We dedicate different MLPs to fit different parts of the image.}
\label{fig:learnINR}
\vspace{-1em}
\end{figure*}

\textbf{Neural Network Spectral Bias. } 
Spectral bias~\cite{rahaman2019spectral,xu2022overview}, or frequency principle~\cite{xu2019frequency,ronen2019convergence}, is a phenomenon that neural networks prioritize learning the low-frequency parts of signals. 
Lots of works have been presented to enable an MLP to fit high-frequency functions, e.g. Fourier feature mapping by Tancik \etal \cite{tancik2020fourier} and periodic activation functions by Sitzmann \etal \cite{sitzmann2020implicit}.  
In this paper, we re-examine the spectral bias and propose the exponential-increase hypothesis from the spatial domain, which is an explicit and quantitative description of the relation between the network convergence rate and the properties of the target signals.

\textbf{Learning-to-learn INRs.} Meta-learning is applied to train a meta-learner that can quickly adapt to new tasks with few training examples. MetaSDF~\cite{sitzmann2020metasdf} first introduced MAML~\cite{finn2017model} to learn excellent INR priors over the respective function space, leading to faster fine-tuning and better geometry reconstruction. Tancik \etal~\cite{tancik2021learned} re-produced such findings with Reptile~\cite{nichol2018first} on a wider variety of signal types.
Yuce \etal~\cite{yuce2022structured} presented a theoretical analysis of meta-learning INRs from the view of dictionary learning.
Based on these works, we show that partition in INR meta-learning framework can modulate the spectral bias within each partition part and improve the effect of learning-to-learn INRs.

\section{Partition for learning and learning-to-learn Implicit Neural
Representations}

\paragraph{Motivations}
Considering a field $\mathbf{q}$ and coordinate $\mathbf{x}$, INR learns a function $\Phi$ with parameters $\Theta$ to fit it, which is denoted as $\mathbf{q}=\Phi(\mathbf{x} ; \Theta)$.
SIREN~\cite{sitzmann2020implicit} shows that MLPs with ReLU activation fail to represent the derivatives of the target signal. So they propose periodic activation functions to represent complex signals and their derivatives. However, even though SIREN is able to represent complex signals, a lot of optimization steps are required due to the fact that too many boundaries with large gradients exist in the spatial domain of the complex signal. We argue that the events of successfully representing each boundary with a large gradient by the neural network parameters $\Theta$ are independent with each other. Then we establish the following hypothesis:

\begin{hypothesis} 
Denote the complexity that one boundary with large derivatives is represented by $\Theta$ as $p$, then the complexity that all boundaries are represented by $\Theta$ is $O(p^N)$, where $N$ is the number of boundaries with large derivatives within the spatial domain.
\label{hypothesis-1}
\end{hypothesis}

We name this hypothesis the exponential-increase hypothesis. Experiments to demonstrate this hypothesis are shown in Appendix~\ref{section:motivation_experiment}. To mitigate the issue caused by this hypothesis, we deliver partition to reduce the exponential complexity of fitting all boundaries to the linear complexity of fitting several regions. Specifically, we divide the whole domain into smaller domains and use independent MLPs to fit a piecewise function to represent the whole function. 

Formally, if the whole domain is divided into $k$ sub-domains, the number of boundaries falling in each sub-domain are $\{N_1, N_2, ..., N_k \}$, where $N = \sum_{i=1}^{k}N_i$. We argue that the optimizations for all MLPs are parallel. If we dedicate neural networks with full capacity to fit each sub-domain, we can assume the complexity of fitting one boundary is still $p$, then the total complexity of parallelly fitting all sub-domains with separate neural networks is
\vspace{-0.3cm}
\begin{equation}
p^{N_1}+p^{N_2}+...+p^{N_k} = \sum_{i=1}^{k} p^{N_i}.
\label{equa:2}
\end{equation}
\vspace{-0.3cm}

Then we can establish the following proposition:

\begin{figure*}[!t]
\centering
\includegraphics[scale=0.73]{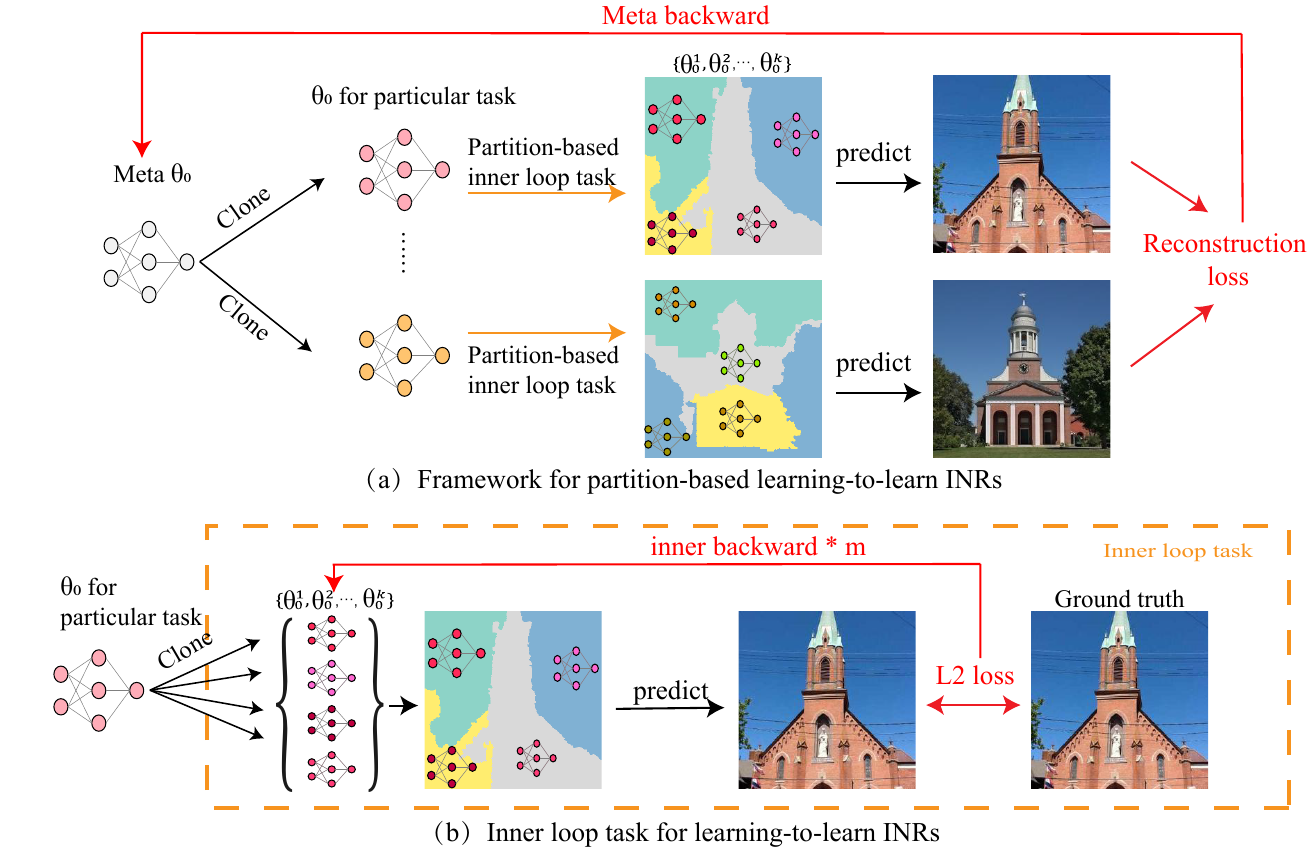}
\caption{(a) Framework for partition-based learning-to-learn INRs. A meta-learner is applied to sample tasks of learning INRs and learns an initialized weight that can quickly fine-tune to a new image. (b) Partition-based inner loop task. The initialized weights will be copied and then optimized for each head within its corresponding sub-domain. (Best view in color.)}
\label{fig:framework1}
\vspace{-1em}
\end{figure*}

\begin{proposition}  
 In case of $k \ge 3$, the complexity of dedicating neural networks with full capacity for each sub-domain is less than the complexity of representing the whole domain with a single neural network, i.e. $\sum_{i=1}^{k} p^{N_i} < p^N$.

\label{proposition-1}
\end{proposition}  

\vspace{-0.5cm}
\label{proof:1}
\begin{proof}

Defining $\hat{N}=\max(N_1,N_2,...,N_k)$, we have:

\vspace{-0.2cm}
\begin{equation}
\frac{\sum_{i=1}^{k}p^{N_i}}{p^{\hat{N}}} = \sum_{i=1}^{k}\frac{p^{N_i}}{p^{\hat{N}}} \le \sum_{i=1}^{k}1 = k .
\label{equa:3}
\end{equation}

Empirically, we should optimize each neural network at least several times, so we have at least $p^{N_i} \ge 2$, then the following inequation holds:
\vspace{-0.2cm}
\begin{equation}
\frac{p^N}{p^{\hat{N}}} = \frac{p^{(N_1+N_2+...+N_k)}}{p^{\hat{N}}} =
\prod_{N_i\neq \hat{N}}p^{N_i} \ge 2^{k-1} .
\label{equa:4}
\end{equation}

\vspace{-0.2cm}
In case of $k\ge 3$, we have $ k < 2^{k-1}$ and $ \sum_{i=1}^{k} p^{N_i} < p^N$. Proposition~\ref{proposition-1} is proved. 
\end{proof}

Theoretically, with larger $k$, Proposition~\ref{proposition-1} can be generalized to the case of fitting each sub-domain with smaller neural networks, whose complexity of fitting one boundary is larger than $p$. We show the proof and the discussion of this case in Appendix~\ref{section:proof1}. Papers about the spectral bias of INRs~\cite{rahaman2019spectral,tancik2020fourier,yuce2022structured} show that INRs are hard to fit the signals with high-frequency components. In fact, the boundaries in the images are high-frequency components of signals and we show that partition helps to reduce the high-frequency components of the input signals in Appendix~\ref{section:high-freq}.

By now, we have mathematically proved that partition can speed up the convergence of INRs by reducing the exponential complexity to linear complexity. And we will present how we practically utilize the partition methods in INRs in the following sections.


\vspace{-1em}
\paragraph{Partition for Learning INRs}
In this part, we show how we can apply partition to learning INRs for 2D images. The framework of the partition-based learning INR method is shown in Figure \ref{fig:learnINR}. We propose to model the INR of a given image $I$ as a weighted sum of $k$ neural networks (denoted as heads). Mathematically, this process can be expressed as 
\vspace{-0.2cm}
\begin{equation}
I(\bold{x})=\sum_{n=1}^k \omega_\phi^n(\bold{x}) I_{\theta_n}(\bold{x}), 
\vspace{-0.2cm}
\end{equation}
where $n$ is the head index. $\omega_\phi^n(\bold{x}): \mathbb{R}^2 \mapsto \{0,1\}$ is the mask for head $n$, and $\omega_\phi( \bold{x}): (\omega_\phi^1( \bold{x}), \omega_\phi^2( \bold{x}),...,\omega_\phi^k( \bold{x})) \in \{0,1\}^k$ is the mask for all heads and satisfies $\left\|\omega_\phi(\mathbf{x})\right\|_1=1$.
This setting ensures that each coordinate in the image is only represented by one single head. 
When predicting an image, each coordinate is only required to be inputted into one single head. Therefore, both the time complexity and memory consumption of predicting the whole image with our partition-based models do not increase.

In practice, we explore two different partition rules for 2D images: one is based on regular grids and the other is based on semantic segmentation maps for 2D images. Detailed implementation is discussed in Section \ref{sec:implementation}.

\begin{figure*}[!th]
\centering
\includegraphics[scale=0.69]{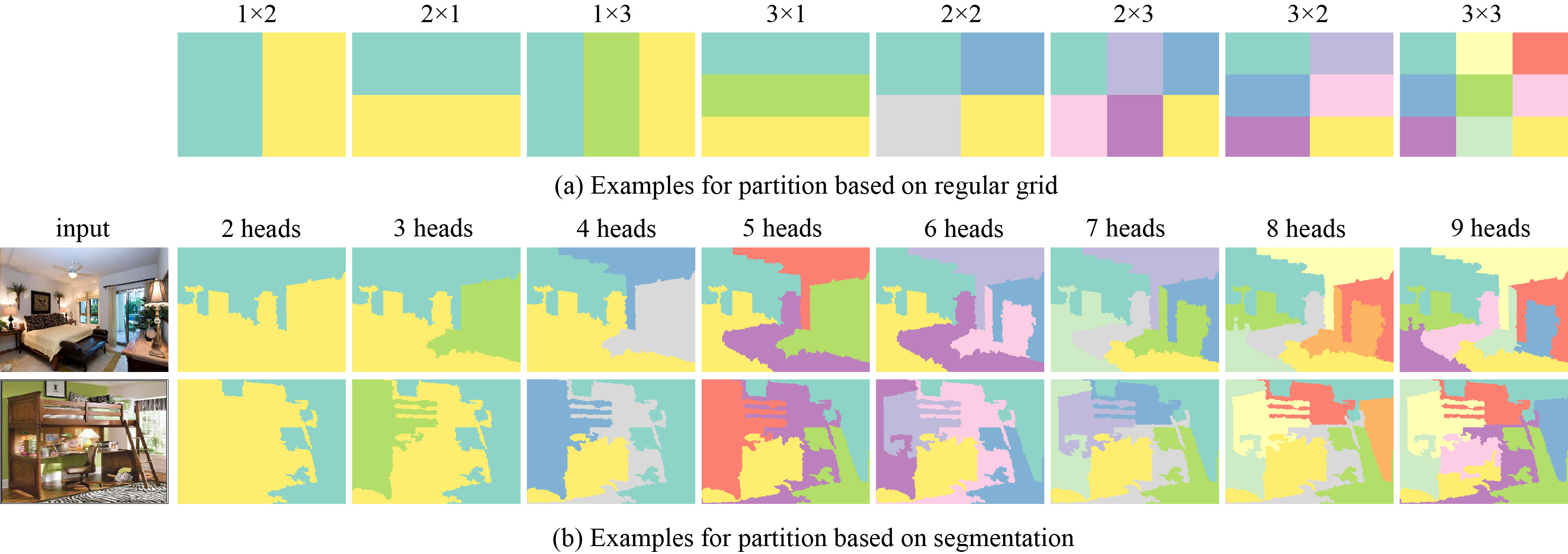}
\caption{(a) Visualization of partition based on regular grids (PoG). (b) Visualization of partition based on HFS semantic segmentation maps (PoS). (Better view in color.)}
\label{fig:decomposition}
\vspace{-0.5em}
\end{figure*}

\vspace{-1em}
\paragraph{Partition for Learning to Learn INRs}
In \cite{sitzmann2020metasdf,tancik2021learned}, they have shown that meta-learning algorithms can provide excellent initial weight parameters for learning INRs, which leads to faster convergence and better generalization. In this part, we show that our partition methods can be integrated into the meta-learning algorithm for INRs, and lead to better generalization and a more flexible inference process than the original meta-learning algorithm for INRs.

Considering a dataset including observations of signals $T$ from a particular distribution $\mathcal{T}$ and a fixed number of optimization steps $m$, the meta-learning algorithms for INRs seek to find an initial weight $\theta_0^*$ that will result in the lowest possible final loss $L\left(\theta_m\right)$ if optimizing a network $f_{\theta}$ for $ m$ steps to represent a new signal from $\mathcal{T}$:  

\vspace{-0.3cm}
\begin{equation}
\theta_0^*=\arg \min _{\theta_0} E_{T \sim \mathcal{T}}\left[L\left(\theta_m\left(\theta_0, T\right)\right)\right].
\vspace{-0.1cm}
\end{equation}

Combining with partition techniques, we partition the whole input domain into $k$ sub-domains with partition rule $\omega$ and seek to find an initial weight $\theta_0^*$ that serves as the initial weight of each head for each sub-domain. This will result in the lowest possible final total loss when optimizing a set of network $F =\{f_{\theta^1}, f_{\theta^2},...,f_{\theta^n},\}$, each of which will represent a part of the new signal from $\mathcal{T}$:

\vspace{-0.5cm}
\begin{equation}
\theta_0^*=\arg \min _{\theta_0} E_{T \sim \mathcal{T}}\left[\sum_{n=1}^kL\left(\theta_{m}^n\left(\theta_0, T, \omega\right)\right)\right].
\end{equation}

We follow MAML~\cite{finn2017model} to learn an initial weight that can serve as a good starting point for gradient descent for all heads. Specifically, given a task $T$ and the number of optimization steps $m$, our partition-based learning-to-learn INRs framework treats these task-specific optimization steps as inner loops, and wraps an outer loop to sample different signals $T_j$ from $\mathcal{T}$. We generate their corresponding partition rules $\omega_j$ to learn the initial weight $\theta_0^*$. Denote the meta-learning rate as $\beta$ and the parameters of head $k$ at $i$ inner loop step and $j$ outer loop step as $(\theta_i^k)_j$, then the updated rule of the parameters is defined as follows: 
\begin{equation}
\vspace{-0.2cm}
(\theta_0)_{j+1}=(\theta_0)_j-\beta \nabla_\theta \sum_{n=1}^kL\left(\theta_m^n\left((\theta_0)_j, T_j, \omega_j\right)\right) .
\vspace{-0.05cm}
\end{equation}

The experiments are conducted in 2D image reconstruction, and direct point-wise observations of the signal $T$ are available. Therefore, we can supervise $F$ with gradient descent using simple L2 loss:
\begin{equation}
L(\theta)=\sum_i\left\|F\left(\mathbf{x}_i\right)-T\left(\mathbf{x}_i\right)\right\|_2^2 .
\end{equation}

So far, we have presented our partition-based learning INRs method as well as the partition-based learning-to-learn INRs method. The architectures of these two methods are presented in Figure \ref{fig:learnINR} and \ref{fig:framework1} respectively.

\section{Implementation}
\label{sec:implementation}
In this section, we introduce two partition rules that both work well under our frameworks. One is based on regular grids (PoG for short) and the other is based on semantic segmentation maps (PoS for short).

\textbf{Partition based on regular grids.} A simple but efficient partition rule is using regular grids to decompose the whole input domain. This method is widely used in image processing tasks based on ViT~\cite{dosovitskiy2020image}, while ~\cite{rebain2021derf,reiser2021kilonerf} have discussed the effect of regular grids decomposition in neural radiance fields tasks. Specifically, for 2D images, we subdivide the input domain into uniform grids of resolution $\mathbf{r} = (r_x,r_y)$, and utilize an independent neural network to fit the content within each grid. Therefore the mapping function $m$ from the pixel position $\mathbf{x}$ to its corresponding neural network index is defined as:
\begin{equation}
m(\mathbf{x})=\Big\lfloor \frac{\mathbf{x}}{\mathbf{r}} \Big\rfloor .
\end{equation}
\textbf{Partition based on Semantic Segmentation Maps.}
We also seek a more flexible and reasonable partition rule, due to the fact that the real images contain non-homogeneous structures and the regular grid partition may violate the continuity of the images. Considering that an in-the-wild image always consists of several parts, it is reasonable to define a sub-domain as the region in which all pixels belong to the same part. Therefore, we seek a partition rule based on image semantic segmentation maps. It is clear that partitioning images based on their semantic segmentation maps helps to reduce the boundaries (or high-frequency components) within each partition part.

Specifically, we start with a hierarchical feature selection (HFS)~\cite{cheng2016hfs} algorithm, which is a rapid image segmentation system and reports over-segmentation results. The over-segmentation results usually assign the regions that are not connected with the same labels and the number of regions is always too large. Thus, we apply the connected-components algorithm on the initial segmentation results to re-label those unconnected parts. And we finally apply a greedy region-merging algorithm to obtain segmentation results with a particular number of regions. 

The performance of PoG and PoS methods are demonstrated in Figure~\ref{fig:decomposition} while we present the formalization of the PoS algorithm in Appendix \ref{section:PoS}.

\vspace{-0.3cm}
\section{Experiments}
In this part, we will first compare the convergence speed of learning the INR for a single in-the-wild image with two modern MLP architectures under the condition of taking partition or not taking partition. We show that our partition methods achieve good performance on both two INR architectures. Then we choose SIREN~\cite{sitzmann2020implicit} as our basic architecture and follow MetaSDF~\cite{sitzmann2020metasdf}'s setting to train meta models with or without partition. 

\begin{figure}[!t]
\centering
\includegraphics[scale=0.36]{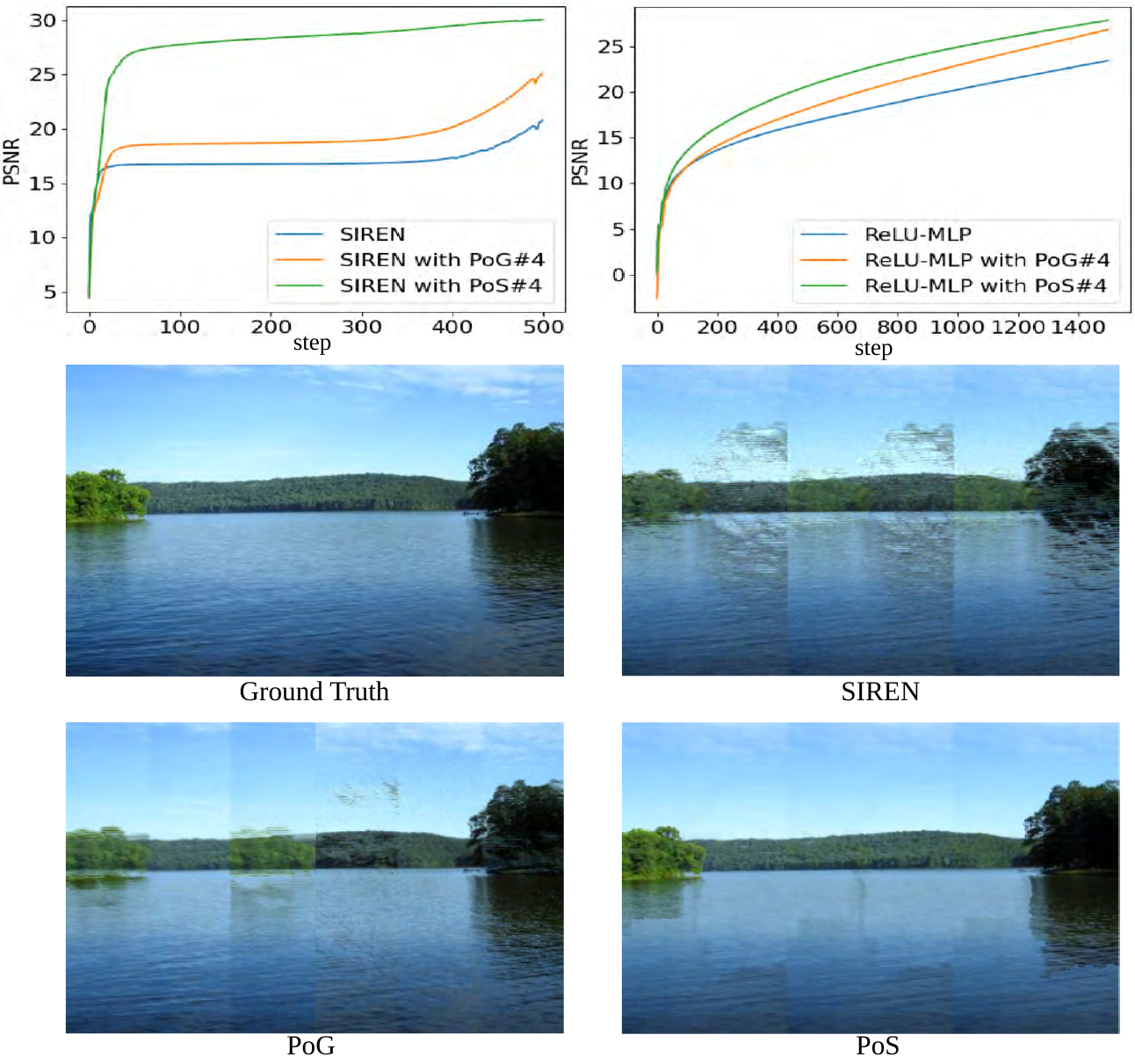}
\caption{The first row presents PSNR vs. step curves for fitting the ground truth image with SIREN-based models and ReLU-MLP-based models (4 heads). The second and third rows present the visual results of optimizing siren-based models for 500 steps. Results from models with partition contain much fewer artifacts with the same optimization steps.}
\label{fig:psnr}
\vspace{-0.3cm}
\end{figure}

\subsection{Partition-based Learning INRs}
\textbf{Settings.} \label{section:setting1}We first choose a landscape image with dimension $380 \times 254$ (shown in Figure \ref{fig:psnr} ground truth) and try to learn an INR for this image. Two popular network architectures are chosen to evaluate our methods: one is SIREN with periodic activation functions~\cite{sitzmann2020implicit} and the other is MLP with ReLU activation functions and positional embedding.

On top of these two baselines, our two partition rules are implemented. To fairly demonstrate the effect of our partition methods, we dedicate neural networks with the same architecture and hyper-parameters but smaller capacity to fit each sub-region, namely heads. We guarantee that the total capacity of all heads is close to the capacity of baselines.
The detailed implementation of these two INR architectures as well as model parameter settings are presented in Appendix \ref{section:detail_architecture}.
Following~\cite{sitzmann2020implicit}'s implementation, we apply the Adam optimizer with a learning rate of $1e-4$. To make a comparison between the two partition methods, the number of partitioned regions is fixed to 4 ($2\times 2$ for PoG).

\begin{figure}[!t]
\centering
\includegraphics[scale=0.5]{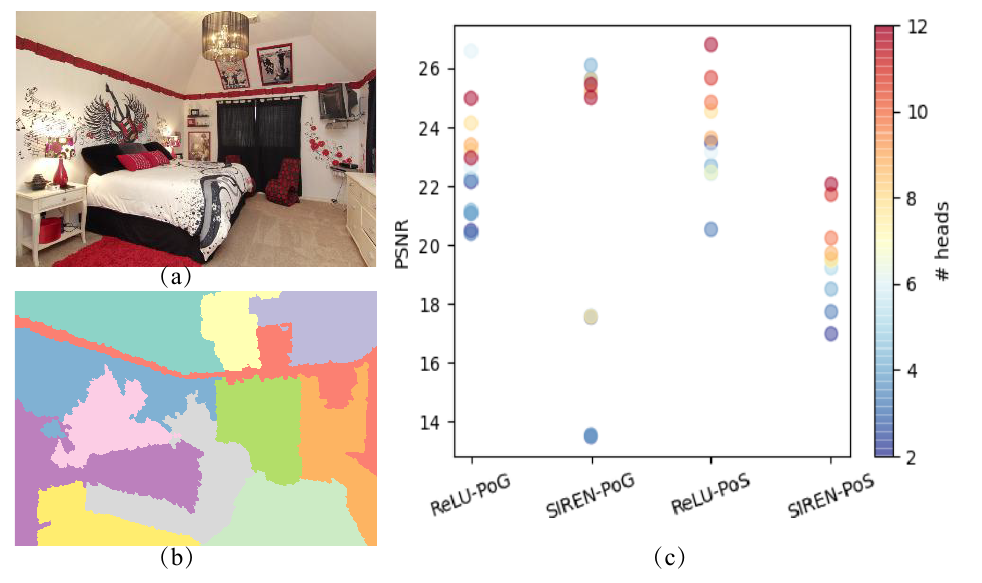}
\caption{Typical results for applying models with different partitioned heads on images. (a) shows the original image and (b) shows its 12-part segmentation result. (c) presents the PSNR values of all models. The results indicate that models with more heads tend to achieve higher PSNR at fixed training steps.}
\label{fig:diff_head_paper}
\vspace{-0.2cm}
\end{figure}



\textbf{Results.} We first present the PSNR curves with respect to optimization steps for applying partition methods on both two baselines, as shown in Figure~\ref{fig:psnr} (the running time and memory consumption of partition-based models are the same or less than the baseline models, which will be discussed in Appendix~\ref{section:more_result_sr}.). On both two baseline architectures, our two partition rules result in faster convergence, while partition based on semantic segmentation maps has better performance than partition based on regular grids. We can observe that for SIREN-based architecture, the model with PoS converges to a high PSNR with very limited steps (less than 100), while the original SIREN needs more than 500 steps to converge to the same PSNR value. For ReLU-MLP-based architecture, the required steps of three cases that the PSNR value reaches 20 are 957, 672, and 445 respectively, which indicates that our partition method based on regular grids (PoG) boosts to $50\%$ speed-up while the partition method based on segmentation maps (PoS) results in $100\%$ speed-up.


\begin{table}[!t]
\begin{center}
\caption{Mean PSNR values for LSUN test images. We optimize the SIREN-based models for 300 steps and ReLU-MLP-based models for 1000 steps. More results are in Appendix~\ref{section:more_result_sr}}  
\begin{tabular}{ll|ll}
\hline
Methods      & PSNRs$\uparrow$ & Methods         & PSNRs$\uparrow$ \\ \hline
SIREN        & 21.211     & ReLU-MLP        & 19.844     \\
SIREN-PoG & 23.864     & ReLU-PoG & 22.672     \\
SIREN-PoS & \textbf{24.485 }    & ReLU-PoS & \textbf{22.863 }    \\ \hline

\vspace{-1.3cm}
\end{tabular}
\label{table:psnr-20}
\end{center}
\end{table}


\textbf{Fewer artifacts for SIREN}. As shown in Figure~\ref{fig:psnr}, SIREN~\cite{sitzmann2020implicit} fails when fitting a large image and tends to generate periodic artifacts. This failure has been reported by \cite{yuce2022structured} and is due to the imperfect frequency recovery. However, the results from our partition-based models generate fewer artifacts with the same optimization steps. This is because partition helps to reduce the high-frequency components that one single SIREN needs to represent, which partly alleviates the negative effect of SIREN.

\textbf{Robustness.} To prove the robustness of our partition methods, we also evaluate our methods on LSUN bedroom image test set~\cite{yu15lsun}, which contains 300 in-the-wild images.
We explore the mean PSNR values with 300 optimization steps for SIREN-based models and 1000 optimization steps for ReLU-MLP-based models. The results are reported in Table \ref{table:psnr-20}. We can observe that both of our partition methods drive the models to a higher PSNR value with the same optimization steps.

\begin{figure}[!t]
\centering
\includegraphics[scale=0.45]{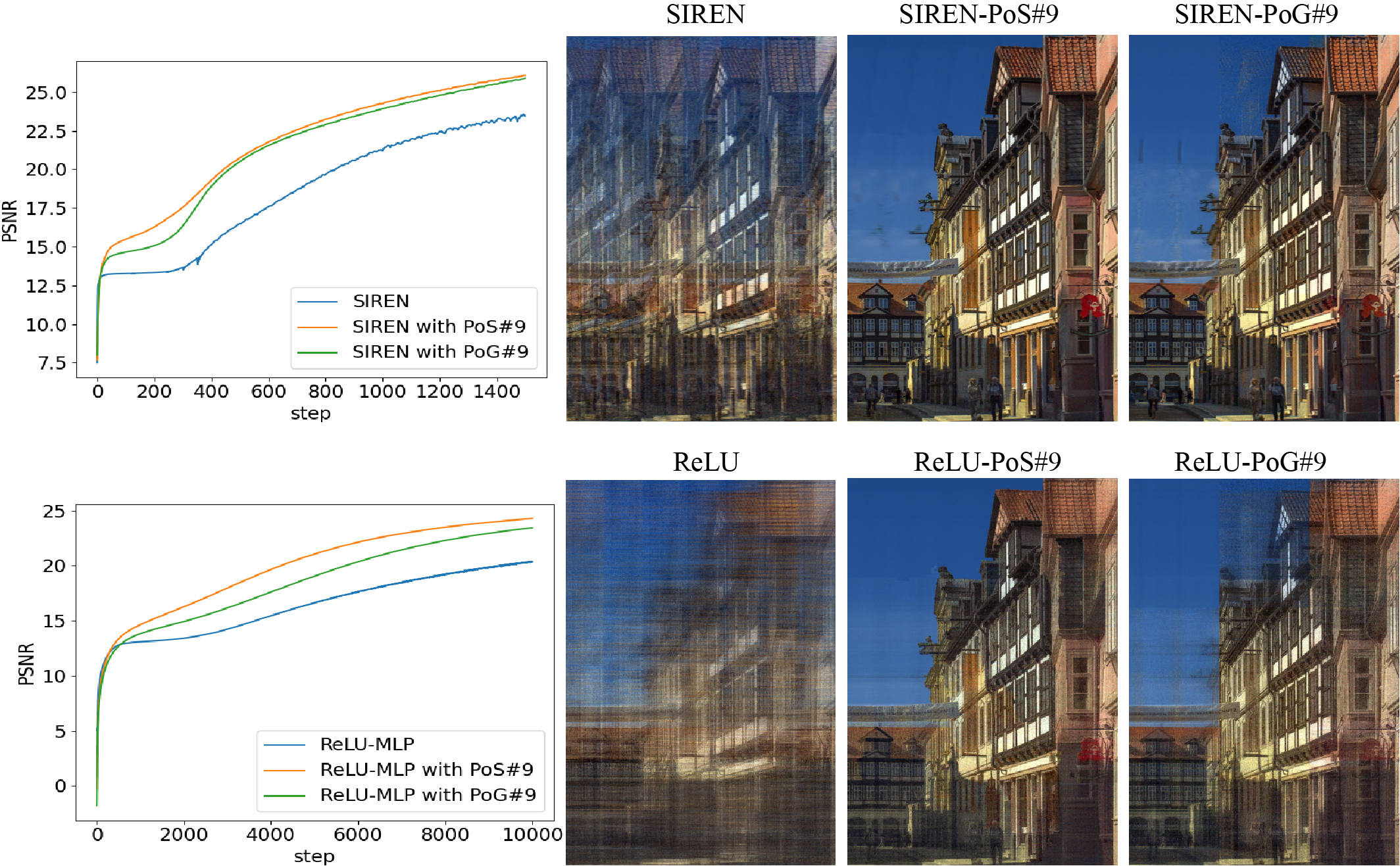}
\caption{Training curves and intermediate results for example super-resolution image. The first row contains the experiments for SIREN-based models while the second row contains experiments for ReLU-MLP-based models. We present the intermediate results of optimizing SIREN-based models for 400 steps and optimizing ReLU-MLP-based models for 4000 steps.}
\label{fig:flickr_results}
\vspace{-0.6cm}
\end{figure}

\textbf{More heads, faster convergence.} We conduct experiments to find the optimal number of heads for both two architectures and two partition methods. A typical example is shown in Figure~\ref{fig:diff_head_paper}. With a different number of heads and two partition methods, we optimize SIREN-based models with 200 steps and ReLU-MLP-based models with 1200 steps. The results show that the models with more heads generally tend to converge to better results and achieve higher PSNR values at the fixed training step. Extensive experiments and discussions of models with different numbers of heads on more images are presented in Appendix~\ref{section: more_experiment_diff_head}.

\textbf{Scale to super-resolution images. } Due to the previous conclusion, we can easily improve the INR optimization efficiency of super-resolution images by increasing the number of partition heads. A typical example of learning INRs for a super-resolution image with $700 \times 1000$ dimension is shown in Figure~\ref{fig:flickr_results}. By partitioning the whole image into 9 parts, we can significantly improve the reconstruction performance of both SIREN-based models and ReLU-MLP-based models. 
More experiments of learning INRs for super-resolution images are presented in Appendix~\ref{section:more_result_sr}.


\subsection{Partition-based Learning-to-learn INRs}
\label{section:setting2}
\textbf{Settings.}  To verify the effect of our partition-based learning-to-learn INRs framework, we follow MetaSDF~\cite{sitzmann2020metasdf} and apply our partition methods in the MAML framework to learn an initial weight that can quickly fine-tune to an unseen image. The outdoor church images from LSUN dataset~\cite{yu15lsun} with the size of $256 \times 256$ are chosen for evaluating the method. The training set contains about 126k images and the test set includes 300 images. Following ~\cite{sitzmann2020metasdf,tancik2021learned}, we choose SIREN as our basic model and set up the number of inner loop step $N$ as 3, which means that our model sees each image only three times. We apply the per-parameter-per-step inner learning rate strategy with initial learning rate $\alpha = 1e-5$. All of the meta-models are trained with an outer loop learning rate $\beta = 1e-4$ and a batch size of 4.

Since our partition methods duplicate the initial weight for each head, we maintain one copy of per-parameter-per-step learning rates for a single head and share it with all heads. The SIREN model in our implementation contains 3 hidden layers and 128 hidden features, which is also the set-up for each head in our models with partition. As a result, the weights trained from the original SIREN and the weights trained from our partitioned-based models have the same keys. Thanks to these settings, in the inference phase we can fine-tune based on our partition methods with the initialized weights trained from the original SIREN.

\begin{table}[!tb]
\begin{center}
\caption{Mean PSNR values performance for SIREN models with different training and fine-tuning methods. For short, we rewrite G as the PoG partition method and S as the PoS partition method. We mark a superscript G/S if we apply the corresponding method in the training stage and mark a subscript G/S if we apply the corresponding method in the fine-tuning stage.}  

\setlength{\tabcolsep}{1.5mm}{
\begin{tabular}{ccc|ccc}
\hline
\multirow{2}{*}{Setting}    & \multicolumn{2}{c|}{PSNR $\uparrow$} & \multirow{2}{*}{Setting}    & \multicolumn{2}{c}{PSNR $\uparrow$} \\
                            & 1 View      & 3 View      &                             & 1 View      & 3 View     \\ \hline
$\text{SIREN \space}$                       & 19.42       & 22.60       & -                           & -           &      -      \\
$\text{SIREN}_{\text{G}}$                    & 19.64       & 22.95       & $\text{SIREN}_{\text{S}}$                    & 19.85       & 23.09      \\
$\text{SIREN}_{\text{G}}^{\text{G}}$ & 19.77       & \textbf{24.23}       & $\text{SIREN}_{\text{S}}^{\text{S}}$ & \textbf{20.00}       & 23.93      \\
$\text{SIREN}_{\text{G}}^{\text{S}}$ & 19.93       & 23.89 & $\text{SIREN}_{\text{S}}^{\text{G}}$ & 18.11       & 20.33     \\ \hline

\end{tabular}
}
\label{table:meta-psnr}
\end{center}
\vspace{-0.7cm}
\end{table}


\begin{figure*}[!t]
\centering
\includegraphics[scale=0.82]{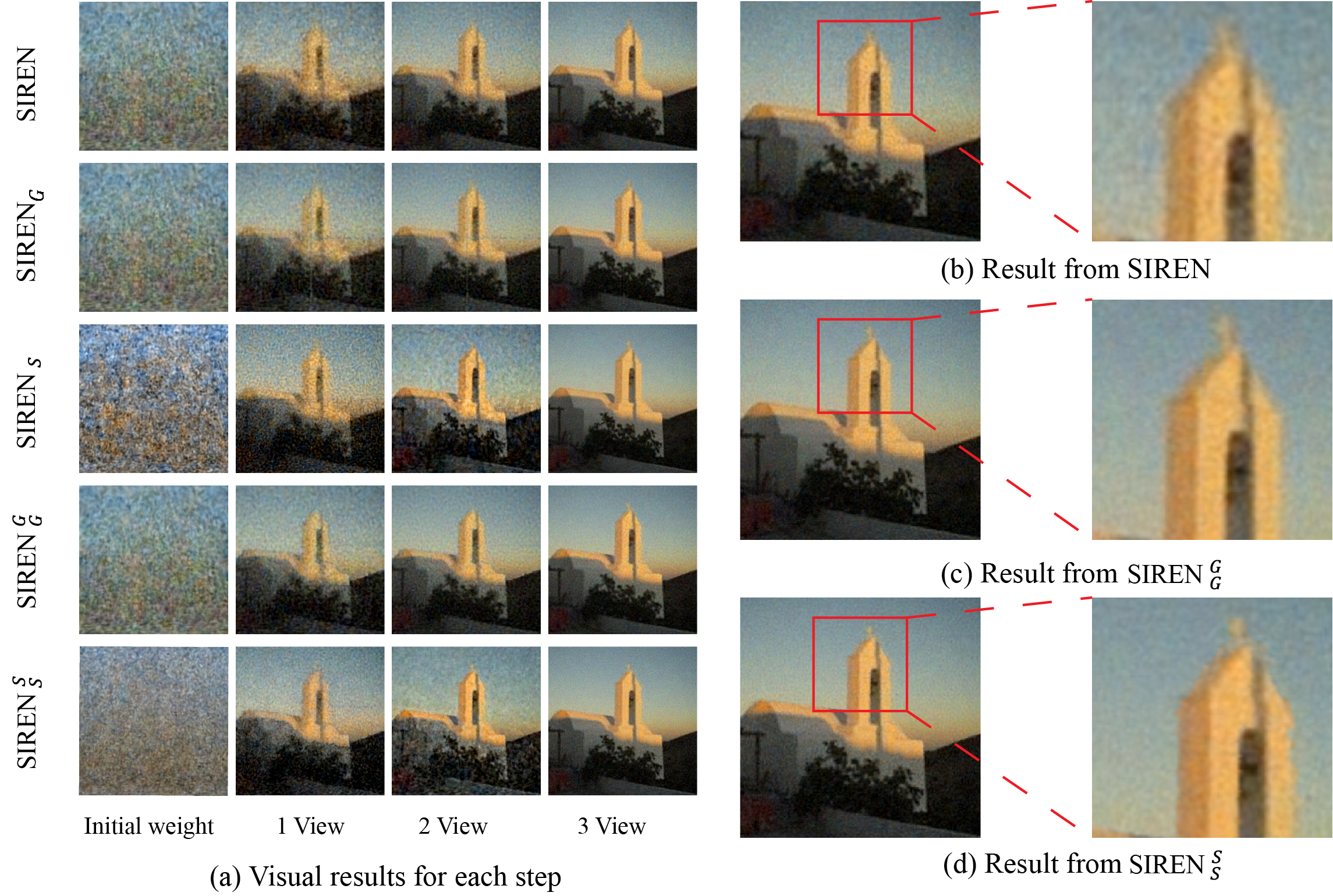}
\caption{Visual results of models with different training and fine-tuning methods. The same abbreviation as in Table \ref{table:meta-psnr}. The results from our partition-based models contain less noise and sharper edges than the SIREN model. (Better view in color.)}
\label{fig:meta-visual}
\vspace{-1em}
\end{figure*}

\begin{figure}[!t]
\centering
\includegraphics[scale=0.85]{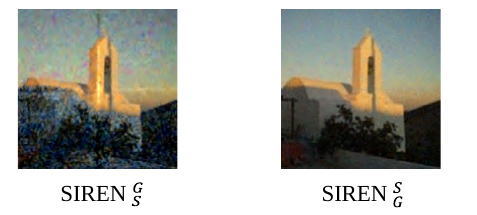}
\caption{Performance of fine-tuning with PoS on the PoG trained initialized weight, and its opposite case. The model trained with PoG but fine-tuned with PoS fails, while the model trained with PoS but fine-tuned with PoG achieves good performance. (Better view in color.)}
\label{fig:meta-visual2}
\vspace{-2em}
\end{figure}

\textbf{Results.} On top of baseline SIREN, We demonstrate the effect of our two partition methods both on the training phase and the inference phase. The mean PSNR values for 1 View and 3 View fine-tuning on 300 images based on models with different training and fine-tuning mechanisms are shown in Table \ref{table:meta-psnr}. The results show that only fine-tuning based on our partition methods with the initialized weights trained from the original SIREN can lead to better performance than baseline, no matter whether we use PoG partition or PoS partition. 
And the experimental results also indicate that the model employing PoG partition during both training and inference phases exhibits the highest PSNR for 3 View fine-tuning, while the model utilizing PoS partition during both phases achieves the highest PSNR for 1 View fine-tuning.
Therefore we prove that fine-tuning based on our partition methods with the initialized weight trained from the partition models has the best performance. For detailed description, a typical example is presented in Figure~\ref{fig:meta-visual}. The result shows that the images obtained from our partition methods contain less noise and sharper boundaries than the result from the baseline SIREN. More visual results and discussions are attached in Appendix \ref{section:more-results}.

\label{section:PoS_PoG}
\textbf{PoS partition as a more flexible choice.} As shown in Figure~\ref{fig:meta-visual2}, fine-tuning with the PoS method from the initialized weights trained with the PoG method leads to a poor result, while the opposite case still maintains good performance. This phenomenon meets our expectations because each head in the PoG method only learns to fit a regular region and fails to fit an irregular region when fine-tuning with the PoS method. On the contrary, the heads in the PoS method learn to fit regions with arbitrary shapes, including the regular grid. As a result, the PoS method can be considered more flexible than the PoG method.

\section{Conclusion}
In this paper, we investigate the dilemma of fitting a discontinuous signal via a continuous function (e.g., a neural network) and demonstrate that the time complexity to force a neural network to fit a discontinuous function is exponentially increasing with the number of high gradients in the input domain, which we call \textit{exponential-increase} hypothesis. We consider the \textit{exponential-increase} hypothesis as a quantitative description of spectral bias~\cite{rahaman2019spectral,tancik2020fourier,yuce2022structured} from the spatial domain. We prove that partitioning the input domain into several sub-domains and dedicating smaller neural networks for each sub-domain help to alleviate this contradiction. Based on this observation, we propose two partition methods for learning and learning-to-learn INRs. We also present two partition rules: one is partitioning based on regular grids and the other is based on semantic segmentation maps. Our methods significantly speed up the convergence of learning INRs from scratch and also lead to better results for fine-tuning a new image at fixed steps for learning-to-learn INRs. Our findings in the paper can serve as theoretical support and inspire the follow-up work on learning more powerful INRs for in-the-wild scenes.


\section*{Acknowledgements}

This work is supported by the National Natural Science Foundation of China (Grant Nos. 62202422 and 61972349), and the National Key Research and Development Program (Grant Nos. 2018YFB1403202, 2019YFF0302601, and 2020YFF0304905). BH is supported by NSFC Young Scientists Fund No. 62006202 and Guangdong Basic and Applied Basic Research Foundation No. 2022A1515011652.

{\small
\bibliographystyle{ieee_fullname}
\bibliography{egbib}
}

\appendix

\clearpage
\section{Experiment for fitting synthetic signals with $N$ boundaries}
\label{section:motivation_experiment}
In this part, we introduce how we generate 1D and 2D synthetic signals and empirically demonstrate the exponential-increase hypothesis. In our synthetic signals, we simplify the boundaries to the place in the input domain where the response values would change from 1 to -1 or from -1 to 1 (it is a special case of large/infinite gradients). Then the methods to generate 1D and 2D signals with specific $N$ boundaries in the spatial domain are as follows:

\begin{itemize}
    \item For 1D synthetic signals with $N$ boundaries, we choose the input domain as [-1,1], and we randomly sample $N$ boundaries. We set the range of $N$ as $[1,70]$. And we randomly sample 5000 points in the input domain as the training set and uniformly sample 5000 points as the testing set.
    \item For 2D synthetic signals with $N$ boundaries, we choose the input domain as $[-1,1]^2$ and we can randomly sample $N_1$ and $N_2$ boundaries in each dimension of the input domain, where the total number of boundaries $N=(N_1+1)\times N_2 + N_1\times(N_2+1) = 2 N_1 N_2+N_1+N_2 = O(N_1N_2)$. To keep the boundaries with uniform distribution along two dimensions, we sample a series of integers $M$ in the range of $[10,250]$ where $N_1, N_2$ as two close factors of $M$ and we calculate the corresponding $N$ based on $N_1, N_2$. We uniformly sample $256 \times 256$ points as the training set and testing set as fitting an image.
\end{itemize}

\begin{figure}[!t]
\centering
\includegraphics[scale=0.55]{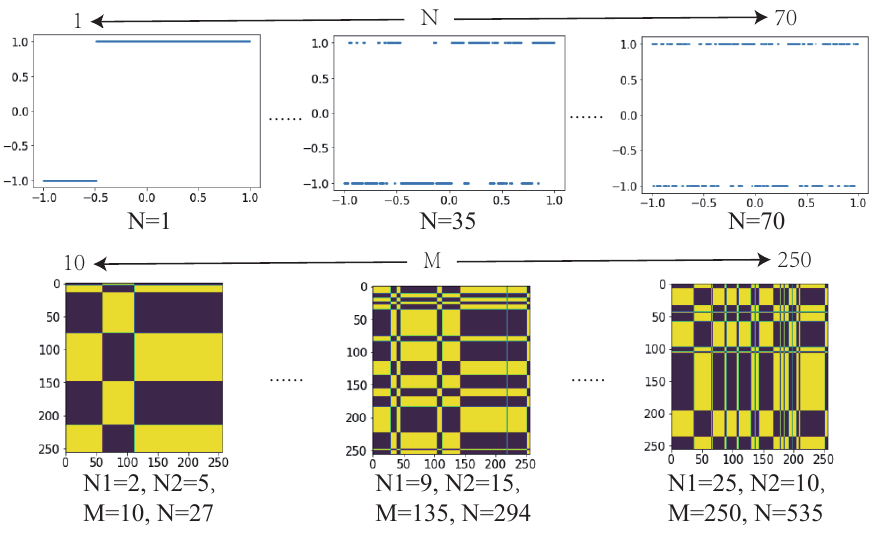}
\caption{Typical examples of the 1D and 2D synthetic signals that we use to demonstrate exponential-increase hypothesis.}
\label{fig:systhsis_signal}
\end{figure}

Typical examples of 1D and 2D synthetic signals with different $N$ boundaries are presented in Figure~\ref{fig:systhsis_signal}.  And the details of delivering SIREN to fit the synthetic signals are as follows. We deliver a SIREN MLP with 32 hidden features and 3 hidden layers to fit the signals. We set $\omega$ for the sinusoidal activation of the first layer for 1D signals as 10 and $\omega$ for the sinusoidal activation of the first layer for 2D signals as 30.  $\omega$ for the sinusoidal activation of the other layer is set to 30.

For both two experiments, MSE loss and the Adam optimizer with a learning rate $1e-3$ are applied to optimize the SIREN models. When the MSE loss is less than 0.05, we think the model fits the signal successfully and report the number of the convergence epoch. The experiments are repeated 5 times and we finally report the mean and std convergence step with respect to the number of boundaries in the spatial domain.  As presented in Figure~\ref{fig:motivation}, the result shows that the curves align with the exponential curves $y=p^N$, where $p=1.0656$ for 1D signals and $p=1.00815$ for 2D signals. Therefore, we empirically demonstrate the exponential-increase hypothesis in learning a single INR.

\section{Complexity of dedicating smaller neural networks}
\label{section:proof1}

In this part, we show that with larger $k$, proposition \ref{proposition-1} can be generalized to the case of fitting each sub-domain with smaller neural networks, whose complexity of fitting one boundary is larger than $p$.

\begin{proof}
Since we dedicate smaller neural networks for each sub-domain, we can assume the complexity of fitting one boundary for each smaller neural network is $\{p_1,p_2,...,p_k \}$, where $p_i > p$. Then the total complexity of parallelly fitting all sub-domain with separate neural networks is 

\vspace{-0.3cm}
\begin{equation}
\vspace{-0.2cm}
p_1^{N_1}+p_2^{N_2}+...+p_k^{N_k} = \sum_{i=1}^kp_i^{N_i}.
\label{equa:10}
\end{equation}

Defining $\hat{N} = \max(N_1,N_2,...,N_k)$ and $\hat{p} = \max(p_1,p_2,...,p_k)$, we hold inequations \ref{equa:11}:

\vspace{-0.5cm}
\begin{align}
\vspace{-0.8cm}
\frac{\sum_{i=1}^kp_i^{N_i}}{\hat{p}^{\hat{N}}} = \sum_{i=1}^{k}\frac{\hat{p}^{N_i}}{\hat{p}^{\hat{N}}}
\le \sum_{i=1}^{k}1 = k
\label{equa:11}
\end{align}

Empirically, we should optimize each neural network several times, so we have $p^{N_i} \ge 2$, then the following inequation hold:
\begin{equation}
\frac{p^N}{\hat{p}^{\hat{N}}} = \frac{p^{(N_1+N_2+...+N_k)}}{\hat{p}^{\hat{N}}} =
\frac{p}{\hat{p}}\prod_{N_i\neq \hat{N}}p^{N_i} \ge \frac{p}{\hat{p}} \cdot 2^{k-1}.
\label{equa:12}
\end{equation}

As long as $k$ is large enough so that $\frac{p}{\hat{p}} \cdot 2^{k-1}> k$, proposition \ref{proposition-1} is proved for smaller neural networks. 
\end{proof}

\vspace{-0.3cm}
\section{Formalization for PoS algorithm}
\label{section:PoS}
The formalization of the PoS algorithm is presented here:

\begin{algorithm}[!ht]
  \caption{Partition Based on Semantic Segmentation Maps }
  \label{alg:hfs}
  \begin{algorithmic}[1]
    \Require
      An image that needed to be partitioned, $I$; the number of required sub-domains $k$
    \Ensure
      The semantic segmentation map, $M$;
    \State Delivering Hierarchical Feature Selection algorithm on $I$ and obtaining initialized segmentation map $M_i$;
    \State Delivering connected components algorithm to re-label those unconnected parts on $M_i$;
    \While{Number of sub-domain($M_i$) \textgreater $ k$ }  
        \State  Finding the sub-domain $D_s$ in $M_i$ which contains smallest area of region;
        \State  Finding the sub-domain $D_r$ who is the neighbor of $D_s$ and contains smallest area of region;
        \State  Changing the label of $D_s$ to the label of $D_r$
    \EndWhile \\
    \Return $M_i$;
  \end{algorithmic}
\end{algorithm}

\section{Partition reduces the high-frequency components of the input signal.}
\label{section:high-freq}

In this part, we show that the partition helps to reduce the high-frequency components that each MLP needs to fit through Fourier Transform. We use PoG as an example but it is clear that this conclusion can be generalized to PoS partition.

\begin{figure}[!h]
\centering
\includegraphics[scale=0.43]{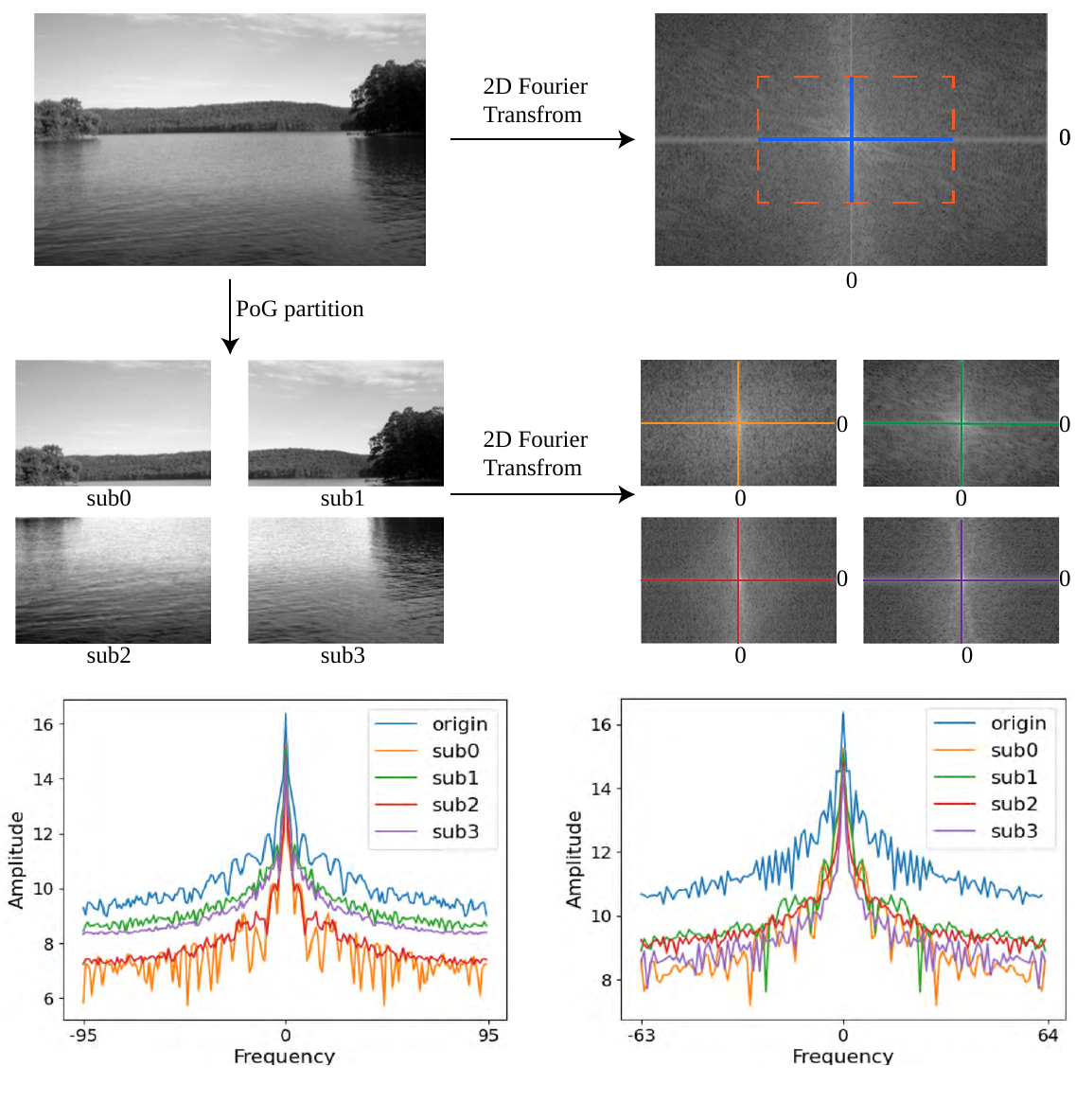}
\vspace{-0.2cm}
\caption{Amplitude analysis of Fourier Transform on image and its sub-parts. We can observe that partition significantly decreases the amplitude of high-frequency components. }
\label{fig:fft_method}
\vspace{-0.5cm}
\end{figure}

Specifically, as shown in Figure~\ref{fig:fft_method}, we apply 2D Fast Fourier Transform~\cite{nussbaumer1981fast} on a given image and shift the zero-frequency components to the center of the plot. We also apply the $2\times 2$ partition on the input image and apply the same Fourier Transform and center-shifting algorithm for each sub-domain. 
We then plot the amplitude vs. frequency for the x dimension while keeping the zero-frequency components at the y dimension, and the amplitude vs. frequency for the y dimension while keeping the zero-frequency components at the x dimension. The experimental results show that the low-frequency components get higher amplitude, which meets the expectation that most information of signals is stored in the low-frequency components. We can also observe that the amplitudes of frequency of all sub-parts are decreased compared with the original image, especially the high-frequency components. From this point of view, we can also prove that our partition helps to improve the convergence speed of learning INRs because it decreases the high-frequency components each MLP needs to fit.

\begin{table*}[!ht]
\begin{center}
\caption{Parameter table for all implemented models. All models contain 3 hidden layers.}  
\vspace*{-5 mm}
\begin{tabular}{cccc}
\\ \hline
\textbf{Base Architecture} & \textbf{\#Hidden Features} & \textbf{\#Head} & \textbf{\#Total Parameters} \\ \hline
SIREN                      & 512                           & 1               & 790017                       \\
SIREN w. partition         & 256                           & 4               & 793604                         \\
ReLU MLPs                  & 512                           & 1               &  911873                           \\
ReLU MLPs w. partition     & 240                           & 4               &  926404                           \\ \hline
\end{tabular}
\label{table:hyper-para}
\end{center}
\vspace{-0.3cm}
\end{table*}

\begin{table*}[!h]
\centering
\caption{Parameter table for models with different heads. We maintain the close total capacity of all models by dedicating smaller hidden features for models with more heads and larger hidden features for models with fewer heads. All heads contain 3 hidden layers.} 
\begin{tabular}{c|cc|cc}
\hline
         & \multicolumn{2}{c|}{SIREN-based architecture} & \multicolumn{2}{c}{ReLU MLP-based architecture} \\ \hline
\# heads & hidden dim           & \# parameters          & hidden dim            & \# parameters           \\ \hline
1        & 512                  & 790017                 & 512                   & 911873                  \\
2        & 360                  & 782642                 & 352                   & 915906                  \\
3        & 296                  & 794763                 & 282                   & 922989                  \\
4        & 256                  & 793604                 & 240                   & 926404                  \\
5        & 228                  & 787745                 & 210                   & 918755                  \\
6        & 208                  & 787494                 & 188                   & 912558                  \\
7        & 192                  & 783559                 & 172                   & 916251                  \\
8        & 180                  & 787688                 & 158                   & 908824                  \\
9        & 170                  & 791019                 & 148                   & 917757                  \\
10       & 162                  & 798670                 & 140                   & 931010                  \\
11       & 154                  & 794497                 & 130                   & 908061                  \\
12       & 146                  & 779652                 & 124                   & 918108                  \\ \hline
\end{tabular}
\label{table:diff_settings}
\vspace{-0.3cm}
\end{table*}

\section{Detailed Experiments Setting for Learning INRs}
\label{section:detail_architecture}

The SIREN \cite{sitzmann2020implicit} and the ReLU MLP with positional embedding are implemented as follows: 

\begin{itemize}
    \item SIREN: we use the sinusoidal function as the activation function for each layer ($y=\sin(\omega x)$), setting up the $\omega$ for the first layer as 60 and the $\omega$ for the hidden layers as 30.
    \item ReLU-MLP with positional embedding: given an input $x$, we use a harmonic embedding layer to convert each feature in $x$ into a series of harmonic features. We set up the number of harmonic functions as 60.
\end{itemize}

To maintain the same capacity of the implementation for models with and without partition techniques, we decrease the number of hidden features of models with partition techniques while maintaining the same number of hidden layers. The detailed parameter setting is presented in Table~\ref{table:hyper-para}.

\begin{figure*}[!h]
\centering
\includegraphics[scale=0.8]{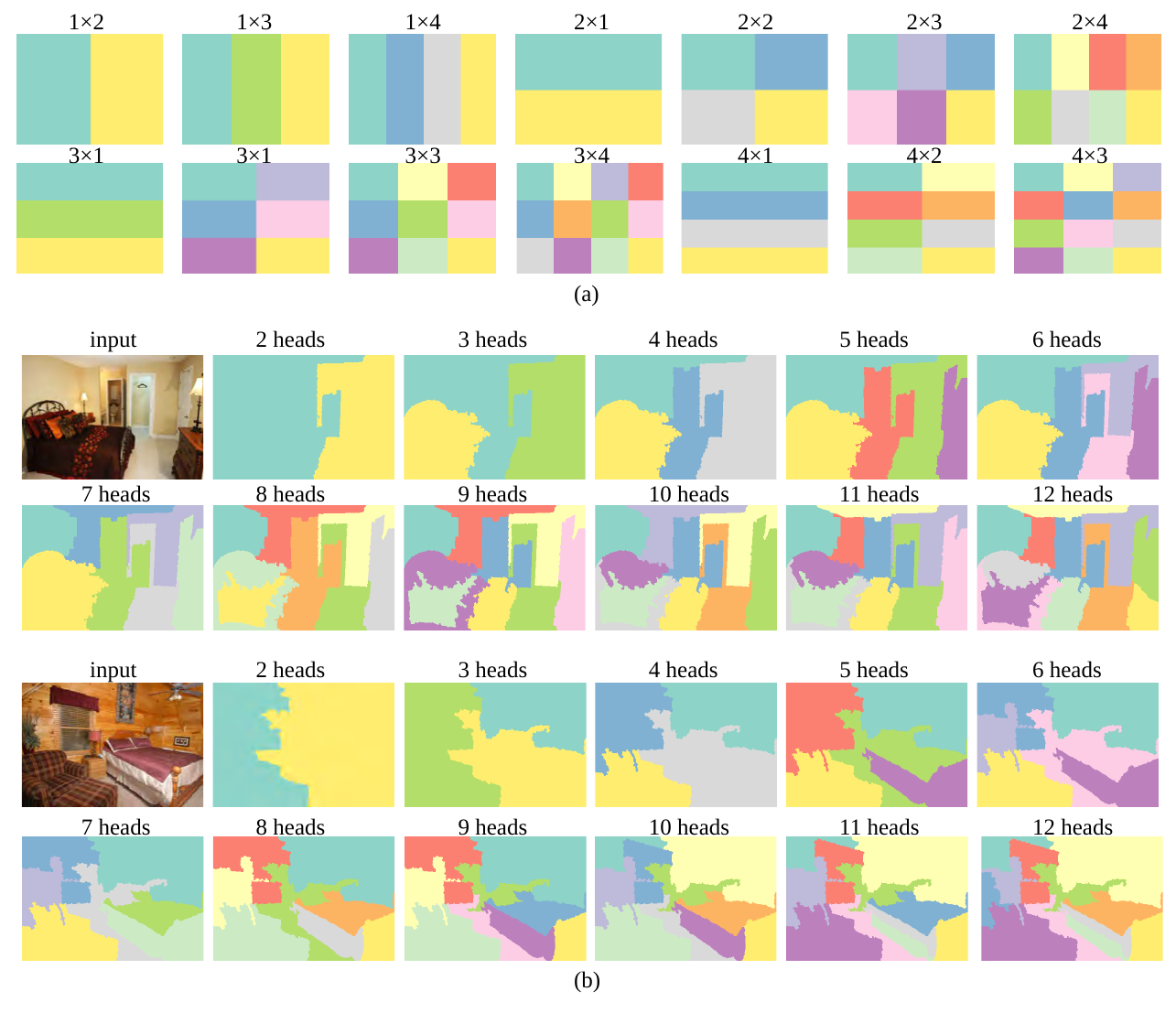}
\vspace{-0.2cm}
\caption{(a) The visual performance of PoG algorithm. We have total 14 cases if we choose $n\in [2,12]$. (b) PoS algorithm performance for two examples (index 7 and 18 in Figure~\ref{fig:diff head}). We can observe that our PoS algorithm can generate good partition masks with balanced area distribution, which is suitable for training piecewise INRs parallelly. }
\label{fig:appendix_PoGS}
\vspace{-0.2cm}
\end{figure*}

\begin{figure*}[!h]
\centering
\includegraphics[scale=0.65]{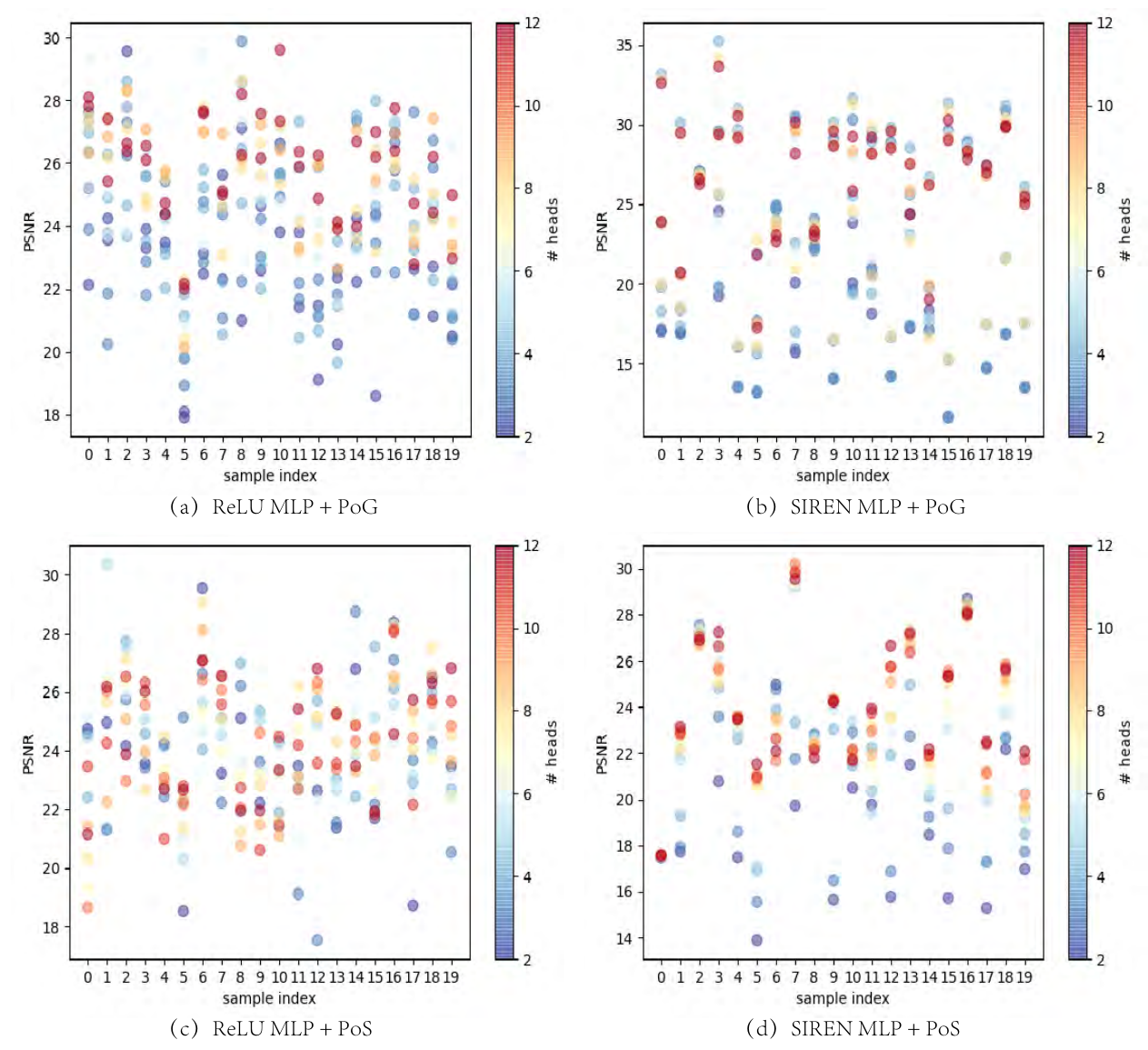}
\vspace{-0.2cm}
\caption{Fixed optimization step performance of models with different heads in 20 LSUN samples. The x-axis denotes which image we are sampling, the y-axis denotes the PSNR for models, and the color denotes the number of heads, which changes from blue to red as the number of heads increases. In both the two architectures and two partition methods, we find that in most samples, red points are mainly concentrated on the top while blue points are mainly concentrated on the bottom. This phenomenon indicates that models with more heads tend to achieve higher PSNR with fixed optimization steps.}
\label{fig:diff head}
\vspace{-0.2cm}
\end{figure*}

\begin{figure*}[!h]
\centering
\includegraphics[scale=0.5]{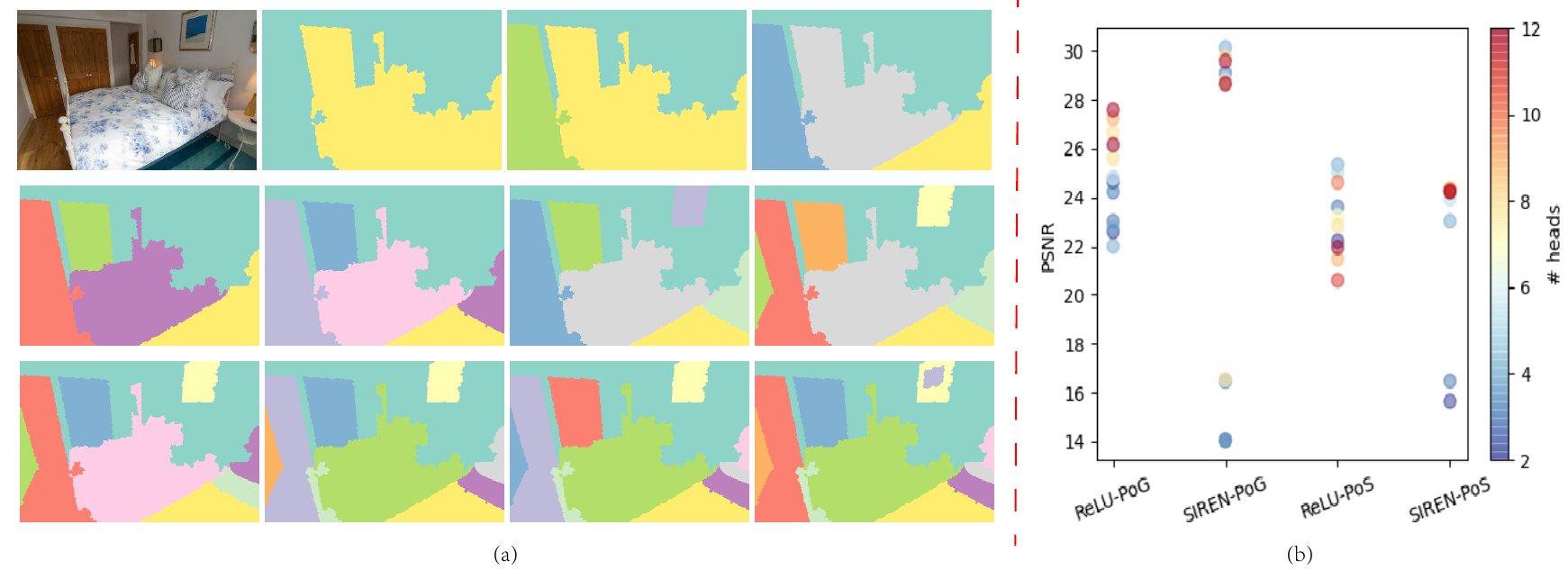}
\caption{A failed case of ReLU-MLP-based models with PoS partition method. (a) The PoS result with $n$ heads, where $n \in [2,12]$. As the number of partition heads increases, the partition area would become more and more imbalanced. When $n=12$, the number of pixels in the largest area is more than 30 times of the number of pixels in the smallest area. (b) In this case, ReLU-MLP-based models with the PoS partition method with more heads fail to achieve higher PSNR values.}
\label{fig:failed_cases}
\vspace{-0.5cm}
\end{figure*}

\section{Experiments for different numbers of heads}
\label{section: more_experiment_diff_head}
In this section, we show the detailed ablation study for the number of heads with our two partition algorithms (PoS and PoG) on two architectures (ReLU MLP and SIREN MLP). We randomly sample 20 images from the LSUN bedroom dataset~\cite{yu15lsun}. For each image, we partition it to $n$ parts, where $n \in [2,12)$, and we dedicate ReLU MLP/SIREN MLP with 3 hidden layers but different hidden features to fit each part. To maintain a close total capacity of all models, we use smaller hidden features for models with more heads and larger hidden features for models with fewer heads. The detailed settings of parameters in all models are presented in Table~\ref{table:diff_settings}. 

Specifically, for PoG we consider partitioning with respect to height and width as two different cases, e.g. $1\times 2$ and $2\times 1$, $1\times 3$ and $3\times 1$, $1\times 4$ and $4\times 1$ are totally 6 different cases. As a result, when  $n \in [2,12]$, PoS has a total of 11 different cases and PoG has a total of 14 cases. In Figure~\ref{fig:appendix_PoGS}, we show more cases for both the PoG algorithm as well as PoS algorithm. We can observe that our PoS algorithm that based on the greedy merging algorithm tends to generate semantic segmentation maps with balanced area distribution within each partition part.

In Figure~\ref{fig:diff head}, we report the PSNR values of optimizing SIREN-based models with 200 steps and ReLU MLP-based modes with 1200 steps. When fitting ReLU-MLP-based models with PoG partition rules, SIREN-based models with PoG partition rules, or SIREN-based models with PoS partition rules, we find that models with more heads tend to achieve higher PSNR values at fixed optimization steps. This phenomenon meets our expectations because the more partition heads we have, the fewer boundaries exist in the area for each head, and the faster convergence for optimizing an MLP with enough capacity. We consider the pixel-based image representation as a special case of our implicit neural representation based on the partition, where the number of partition heads increases to the number of total pixels. Under this condition, each head is represented by a constant function. The heads in this special case only need a few optimization steps.

When fitting ReLU-MLP-based models with PoS partition rules, we observe a lot of failed cases. Some models with more partition heads converge to lower PSNR values at fixed optimization steps. A typical example is shown in Figure~\ref{fig:failed_cases}. We think this is due to the imbalanced partition areas of the PoS method and the limited expressive power of ReLU MLP. If we make an average distribution of the limited capacity of ReLU MLP over all heads, the capacity of a single ReLU MLP head is not enough to fit a large area and is over-fitting for a small area. As the number of heads increases, the difference between the largest area and the smallest area will become larger, which ultimately leads to the failure of fitting ReLU-MLP-based models with PoS partition rules over too many heads. Indeed, we can alleviate this problem by adaptively allocating the capacity of ReLU MLP heads according to the area they would fit.

\section{More results of learning INRs.}
\label{section:more_result_sr}
In this part, we show that we can easily improve the convergence speed of learning INRs for super-resolution images by dividing the whole image into more parts. In addition, we show that our algorithm can also speed up the training of shape regression and NeRF.

\begin{figure*}[!h]
\centering
\includegraphics[scale=0.42]{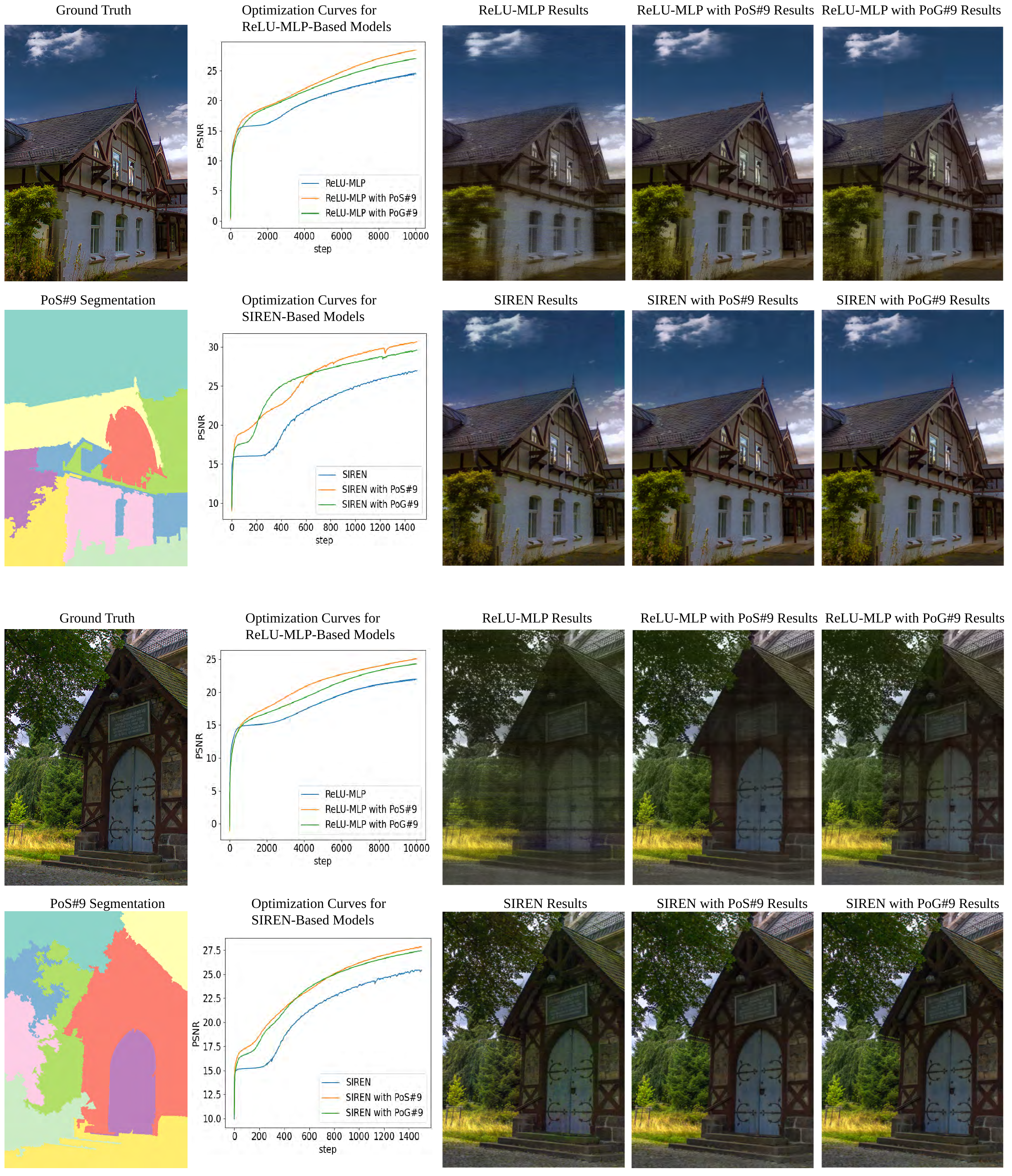}
\vspace{-0.2cm}
\caption{Performance of our partition-based models in two examples of super-resolution images. We present the ground truth, PoS segmentation maps with 9 parts, the optimization curves as well as the reconstruction results of all models (10000 steps for ReLU-MLP-based models and 1500 steps for SIREN-based models). We observe that our partition-based models lead to better reconstruction results (about 2.5 to 5 dB higher PSNR than baselines). We can also observe that the baselines fail to reconstruct the color of the tree in the lower left corner of the first image and the color of the door in the second image. All models have the same capacity. }
\label{fig:super_more_results}
\vspace{-0.5cm}
\end{figure*}

\textbf{Results for images from Flickr1024 dataset.} As shown in Figure~\ref{fig:super_more_results}, we choose an image with dimension $700 \times 1200 $ and an image with $700 \times 1100$ from Flickr1024 dataset. We apply models with 9 partitions heads (the number of hidden layers and the hidden features are the same as the model with 9 heads presented in Table~\ref{table:diff_settings}) to learn INRs for both images ($3\times 3$ for PoG, and segmentation maps with 9 heads for each image are shown in Figure~\ref{fig:super_more_results}). We optimize the SIREN-based models for 1500 steps and ReLU-MLP-based models for 10000 steps, for the sake of getting a good visual performance comparison.

From the optimization curves in Figure~\ref{fig:super_more_results}, we can find that the results from the partition-based models have about $2.5 - 5$ dB higher PSNR than the baseline models. From the visual results, we can observe that the results reconstructed from the baselines contain much more noise. And the baselines fail to reconstruct some parts of the images, e.g. the color of the tree in the lower left corner of the first image and the color of the door in the second image.

\textbf{Less running time and memory consumption for our partition-based models.} We also report the total running time and memory consumption of our partition-based models and the baseline models on the two Flickr1024 images described before. The results are presented in Table~\ref{table:time_memory}. It is evident that our partition-based models have significantly reduced running time and GPU memory consumption compared to the baseline models while having the same capacity. This observation demonstrates that our partition techniques not only accelerate the learning process of INRs algorithmically but also in practical terms.

\begin{table*}[!t]
\caption{Running time and memory consumption for our partition-based model and the baseline models with the same parameter capacity. We optimize the SIREN-based models for 1500 steps and ReLU-MLP-based models for 10000 steps. } 
\begin{tabular}{c|ccc|ccc}
\hline
Image size                & Model        & Time (min) & Memory (GB) & Model           & Time (min) & Memory Cost(GB) \\ \hline
\multirow{3}{*}{700*1200} & SIREN        & 5.7        & 18.9        & ReLU MLP        & 32.1       & 15.8            \\
                          & SIREN PoS\#9 & \textbf{2.5}        & 7.6         & ReLU MLP PoS\#9 & \textbf{20.1}       & 6.5             \\
                          & SIREN PoG\#9 & 2.9        & \textbf{7.2}         & ReLU MLP PoG\#9 & 20.3       & \textbf{5.8}             \\ \hline
\multirow{3}{*}{700*1100} & SIREN        & 5.2        & 17.5        & ReLU MLP        & 29.5       & 14.7            \\
                          & SIREN PoS\#9 & \textbf{2.7}        & 7.0         & ReLU MLP PoS\#9 & \textbf{17.5}       & 6.2             \\
                          & SIREN PoG\#9 & 2.8        & \textbf{6.7}         & ReLU MLP PoG\#9 & 17.8       & \textbf{5.7}             \\ \hline
\end{tabular}
\label{table:time_memory}
\end{table*}

\textbf{Scale to $8000 \times 8000$ image.} 
Same as presented in ACORN\cite{martel2021acorn}, we present an experiment to fit a dwarf planet Pluto image with $8000 \times 8000$ dimension.   Please note that rather than making a performance comparison with ACRON\cite{martel2021acorn}, our target is to show that based on the exponential-increase hypothesis, simply partitioning the input domain and delivering smaller MLPs for each region can speed up the convergence rate of fitting an extremely large image. The fast training and inference efficiency of ACORN should give the credit to its hybrid multiscale architecture but not the partition strategy. 


\begin{figure}[!t]
\centering
\includegraphics[scale=0.55]{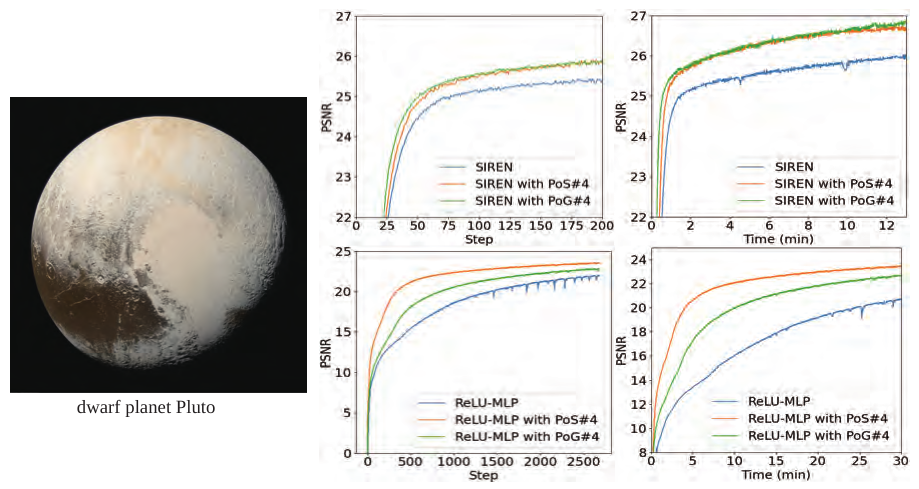}
\vspace{-0.2cm}
\caption{The left image is the dwarf planet Pluto with $8000\times 8000$ pixels. The fitting experiment results on the right show that our partition methods can also speed up the convergence rate of learning INRs for extremely large images.}
\label{fig:pluto}
\vspace{-0.5cm}
\end{figure}

The experimental settings are as follows. Due to the large resolution of the dwarf planet Pluto image, it is impossible to input all pixels into the models within one batch, like the training process in the experiments presented in section~\ref{section:setting1}. Therefore, we use the mini-batch training strategy to train the models. More specifically, within a single batch, 500,000 pixels are sampled randomly to train the models. Since the dwarf planet Pluto images does not have too many boundaries, we partition it into 4 sub-regions and dedicate 4 smaller SIREN MLPs with 758 hidden features and 4 hidden layers to fit it. The performances of our partition-based models are compared with the baseline single SIREN MLP with 1536 hidden features and 4 hidden layers, which has the same capacity as our partition-based models. Other hyper-parameters are the same as the experiments in section~\ref{section:setting1}.

We report the fitting PSNR with respect to the training epoch as well as training time in Figure~\ref{fig:pluto}. The results show that the PSNR values at the same step or training time for our partition-based models are much higher than the baseline models. Therefore, our partition methods work well for extremely large images.

\begin{figure}[!t]
\centering
\includegraphics[scale=0.7]{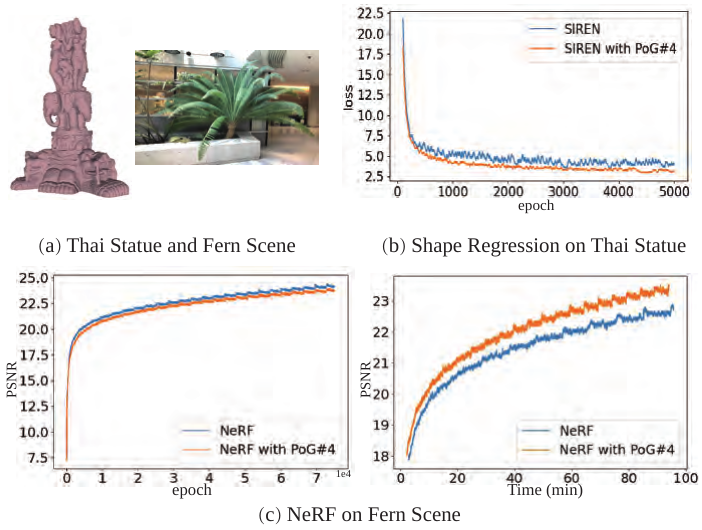}
\vspace{-0.2cm}
\caption{We apply partition-based INR learning algorithm on Thai Statue (point cloud) and Fern Scene (NeRF). The results show that our algorithm can speed up the learning for 3D data.}
\label{fig:shape&NeRF}
\vspace{-0.5cm}
\end{figure}

\begin{table*}[h]
\centering
\caption{Times of best 3 Views performance on 300 images for all models.} 
\begin{tabular}{c|ccccccc}
\hline
\textbf{Architectures}       & $\text{SIREN}_{\text{G}}^{\text{G}}$ & $\text{SIREN}_{\text{S}}^{\text{S}}$ & $\text{SIREN}_{\text{G}}^{\text{S}}$ &
$\text{SIREN}_{\text{G}}$ & $\text{SIREN}_{\text{S}}$ & SIREN & $\text{SIREN}_{\text{S}}^{\text{G}}$ \\ \hline
\textbf{\# highest PSNR} & 273                         & 17                          & 0                           & 0        & 0        & 0     & 0                           \\ \hline
\end{tabular}
\label{table:times}
\end{table*}

\textbf{Scale to Shape Regression and NeRF.} To demonstrate the performance of partition, we also test our algorithm in 3D data, including shape regression on Thai Statue and NeRF on Fern Scene, as presented in Figure~\ref{fig:shape&NeRF} (a). 

To regress the shape on Thai Statue, we follow SIREN~\cite{sitzmann2020implicit} and sample on\_surface\_coords and off\_surface\_coords from Thai Statue point cloud. The signed distance functions (SDFs) of on\_surface\_coords are assigned to 0 while the SDFs of off\_surface\_coords are assigned to -1. We apply SIREN and SIREN with PoG with 4 heads to fit the data. The partition mask is based on grid that has boundaries on the mean of the x dimension and y dimension. We show the training curve in Figure~\ref{fig:shape&NeRF} (b), which shows that partition leads shape regression to lower errors and smaller oscillation.

For NeRF on Fern Scene, we also compare the baseline NeRF~\cite{mildenhall2021nerf} with our partition-based NeRF. More specifically, we apply a partition mask with $2\times 2$ regular grids that have boundaries on the mean of the x dimension and y dimension. The results show that our partition speeds up the training of NeRF in terms of training time but not training epoch. We argue that the phenomenon occurs because the calculation of each pixel value requires the evaluation of coordinates that are located in different partition domain, which makes the partition mask blurred.

\section{More results and discussion for Learning-to-learn INRs}
\label{section:more-results}

We explore which model achieves the highest 3 Views PSNR value on all 300 images and present the results in Table \ref{table:times}. We find that the $\text{SIREN}_{\text{G}}^{\text{G}}$ model achieves best performance in most cases, while  $\text{SIREN}_{\text{S}}^{\text{S}}$ achieves best performance in the rest cases. To some extent, this result violates our expectations since the PoS has a better performance than PoG when training an INR for a single image. As discussed in section~\ref{section:PoS_PoG}, PoS could be considered as a more general choice than PoG. We think the reason why PoS fails to get better performance than PoG on the learning-to-learn INRs framework is that training a good initialized weight that is suitable for the arbitrary shape of sub-domain (PoS) is much harder than training a good initialized weight that is suitable for the regular grid (PoG).

\begin{table}[!t]\footnotesize
\centering
\setlength\tabcolsep{3.5pt}
\caption{\footnotesize Mean SSIM and  LPIPS for LSUN test images.}
\begin{tabular}{lll|lll}
\hline
Methods   & SSIM$\uparrow$                    & LPIPS$\downarrow$ & Methods  & SSIM$\uparrow$                    & LPIPS $\downarrow$ \\ \hline
SIREN     & 0.744                            &        0.343         & ReLU & 0.415                          & 0.806                               \\
SIREN-PoG & 0.839                            &      0.233           & ReLU-PoG & 0.430                            &      0.743                          \\
SIREN-PoS & \textbf{0.862} &     \textbf{0.197}            & ReLU-PoS & \textbf{0.461} &  \textbf{0.660}                              \\ \hline
\end{tabular}
\label{table:ssim}
\end{table}

\begin{figure*}[!h]
\centering
\includegraphics[scale=0.53]{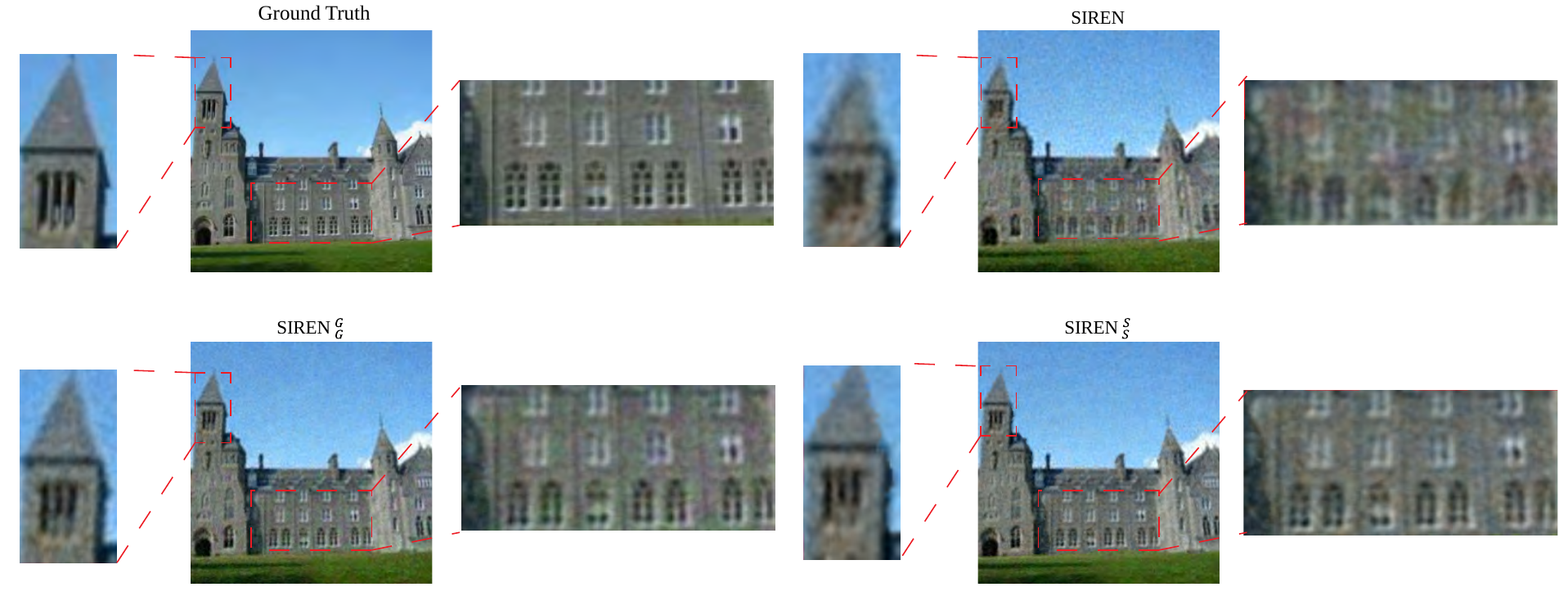}
\vspace{-0.2cm}
\caption{An example of performance comparison of 3 Views results of the baseline model (SIREN) and our partition-based models. Our partition-based model would generate better results that contain less noise as well as sharper boundaries.}
\label{fig:more-meta-visual2}
\vspace{-0.5cm}
\end{figure*}

\begin{figure*}[!h]
\centering
\includegraphics[scale=0.65]{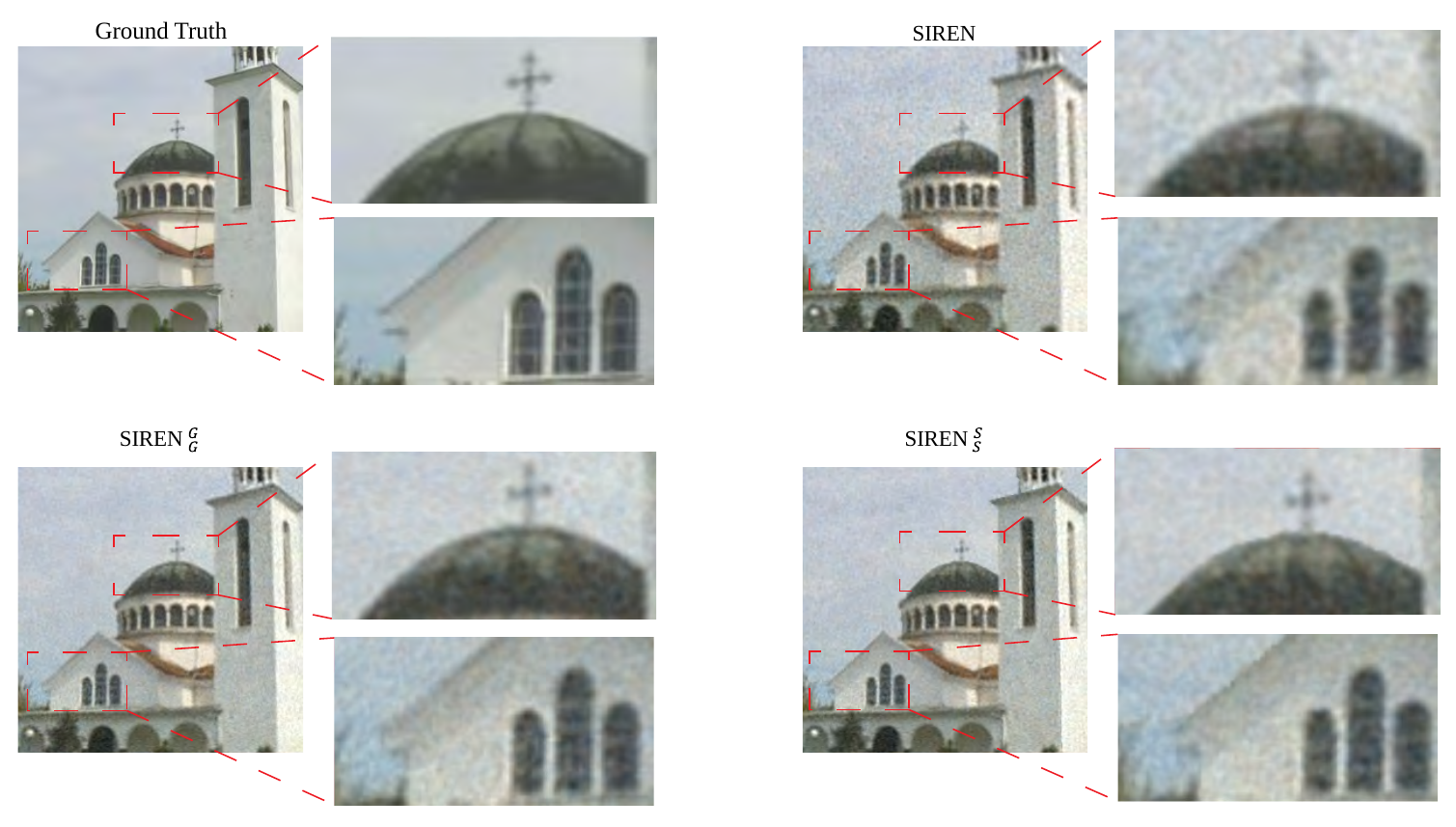}
\vspace{-0.2cm}
\caption{Another example of performance comparison of 3 Views results of the baseline model (SIREN) and our partition-based models. Our partition-based model would generate better results that contain less noise as well as sharper boundaries.}
\label{fig:more-meta-visual1_2_2}
\vspace{-0.5cm}
\end{figure*}

\begin{figure*}[!h]
\centering
\includegraphics[scale=0.75]{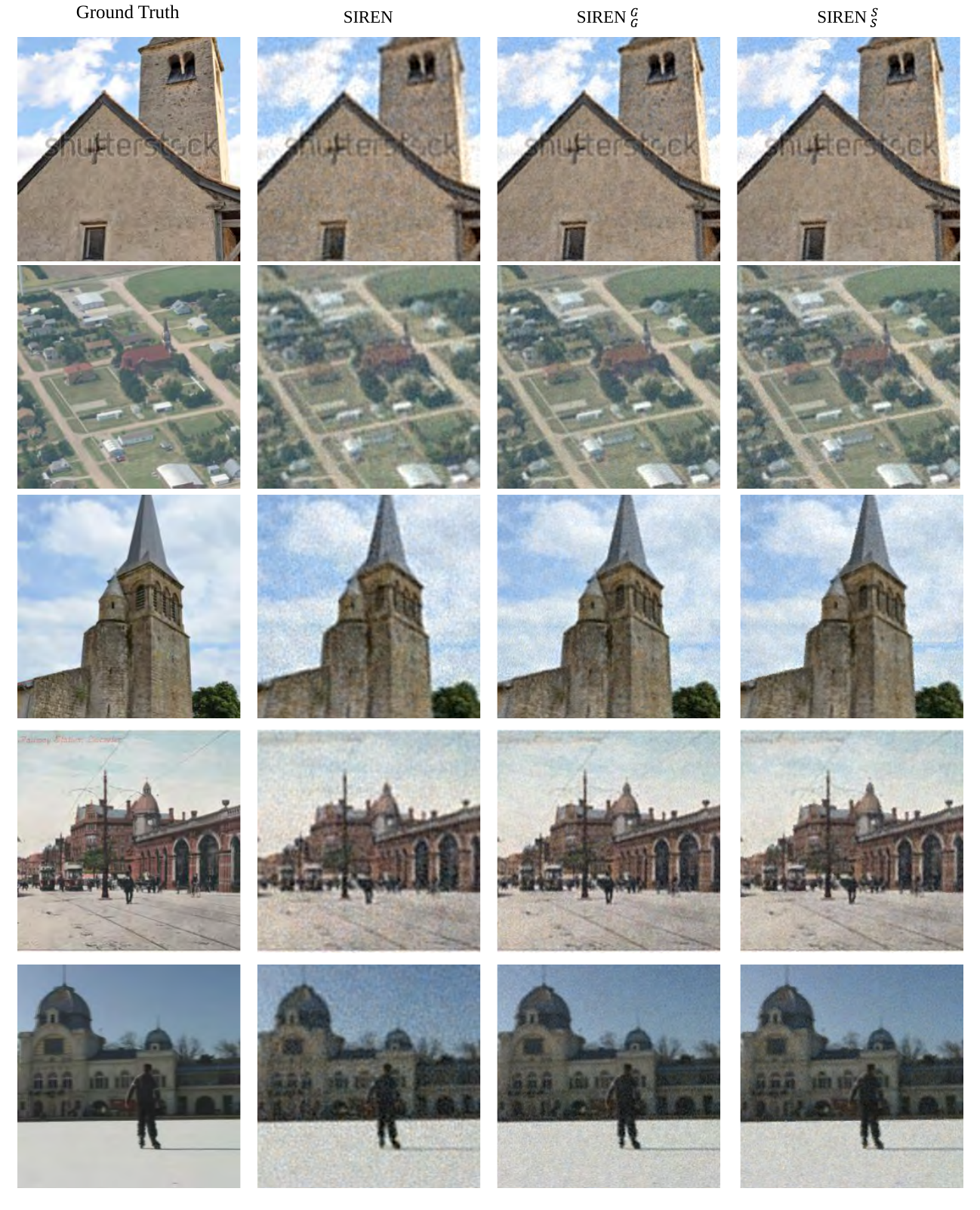}
\vspace{-0.2cm}
\caption{More examples of performance comparison of 3 Views results of baseline model and our partition-based models. Our partition-based model would generate less noise as well as sharper boundaries.}
\label{fig:more-meta-visual2_2}
\vspace{-0.5cm}
\end{figure*}

We present more examples that our partition-based models defeat the baseline SIREN in Figure \ref{fig:more-meta-visual2}, \ref{fig:more-meta-visual1_2_2} and \ref{fig:more-meta-visual2_2}. We can see that our partition-based models generate less noise than the baseline models. And the partition-based models also generate sharper discontinuous boundaries between two separate objects in the scene, which further proves our Proposition \ref{proposition-1}.

\begin{figure*}[!h]
\centering
\includegraphics[scale=0.62]{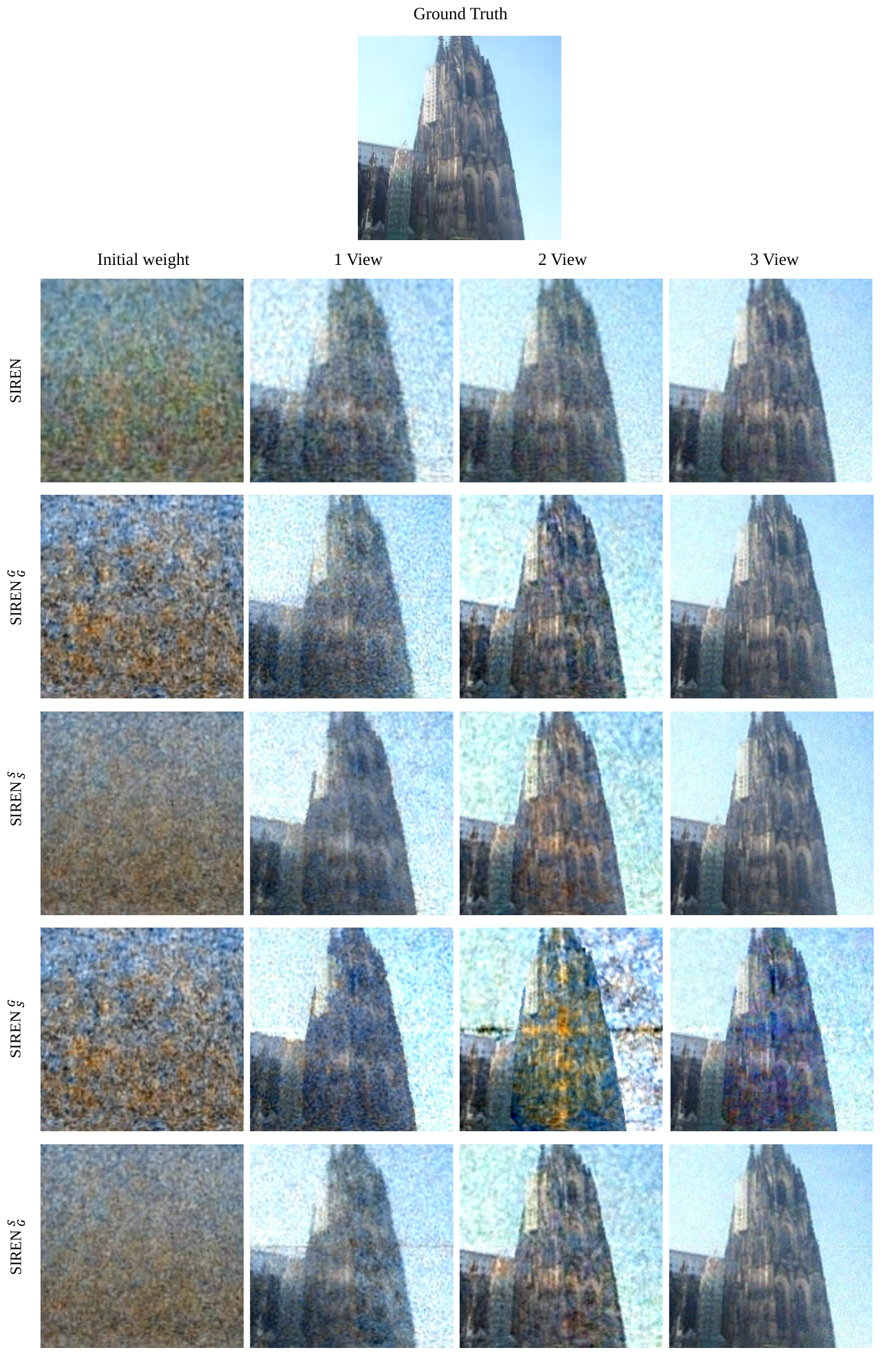}
\vspace{-0.2cm}
\caption{Example 1 of performance comparison of results of models with different training and fine-tuning mechanisms. We can observe that the model trained with PoG and fine-tuning with PoS fails but the model trained with PoS and fine-tuning with PoG achieves good performance.}
\label{fig:more-meta-visual3}
\vspace{-0.5cm}
\end{figure*}

\begin{figure*}[!h]
\centering
\includegraphics[scale=0.62]{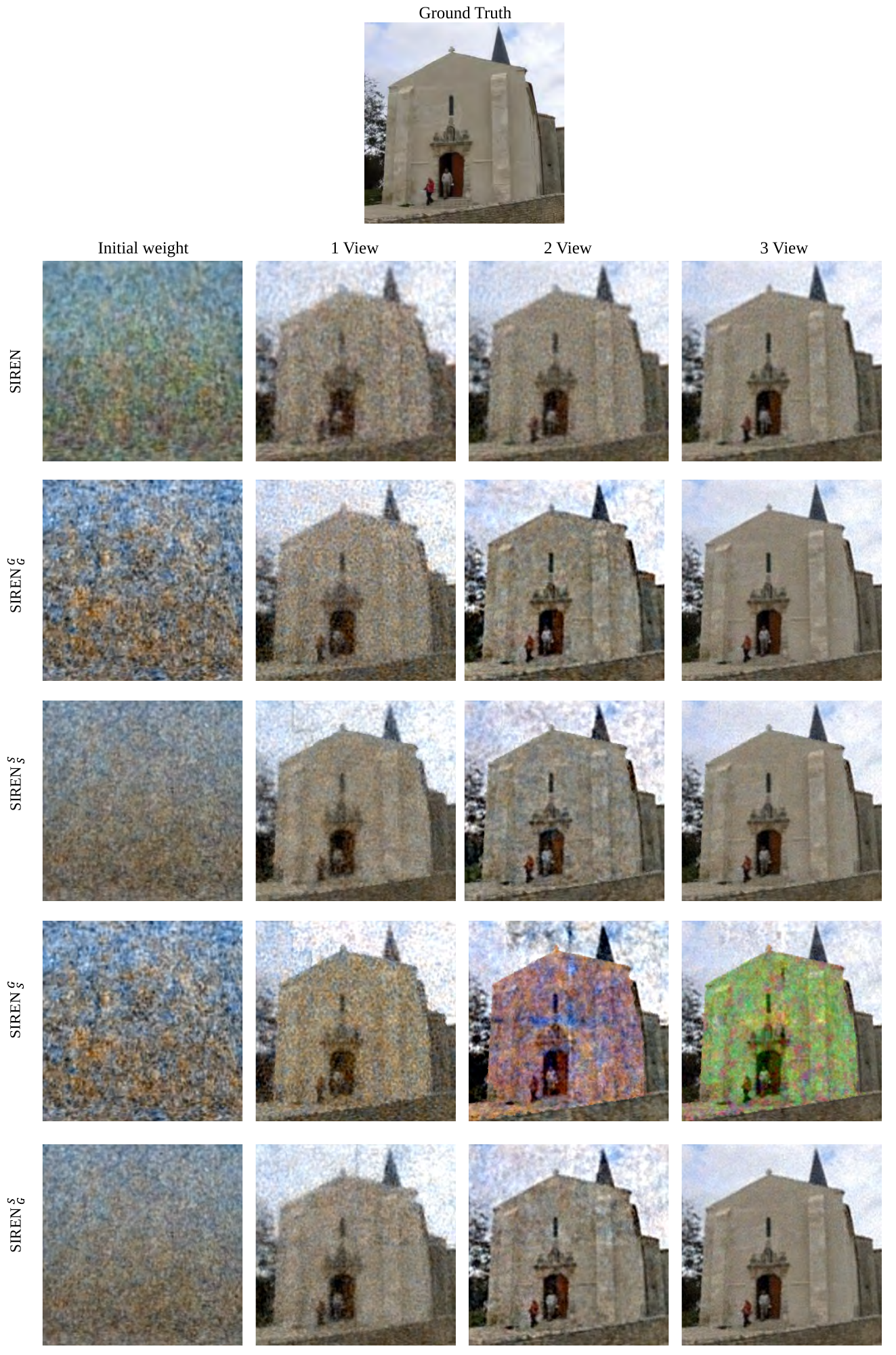}
\vspace{-0.2cm}
\caption{Example 2 of performance comparison of results of models with different training and fine-tuning mechanisms. We can observe that the model trained with PoG and fine-tuning with PoS fails but the model trained with PoS and fine-tuning with PoG achieves good performance.}
\label{fig:more-meta-visual4}
\vspace{-0.5cm}
\end{figure*}

In Figure~\ref{fig:more-meta-visual3} and ~\ref{fig:more-meta-visual4}, we also show two more examples of the detailed fine-tuning process of models with a different setting.  We can observe that the model trained with PoG and fine-tuning with PoS fails but the model trained with PoS and fine-tuning with PoG achieves good performance. This phenomenon gives evidence that PoS is more flexible than PoG.

\section{More quantitative results}
\label{section:more-quantatitative-results}
In this section, we provide the additional quantitative results in terms of two more metrices, SSIM~\cite{wang2004image} and LPIPS~\cite{zhang2018unreasonable}. 
Distinguished from PSNR that only measures absolute error, SSIM is a perception-based model that considers image degradation as perceived change in structural information while LPIPS delivers deep neural networks to provide an emergent embedding which agrees well with humans.

Table~\ref{table:ssim} provides supplemental results for Table~\ref{table:psnr-20}. The result agrees well with Table~\ref{table:psnr-20} and shows that our partition indeed leads learning INRs methods to better reconstruction performance from the human's perception.
Table~\ref{table:ssim2} provides supplemental results for Table~\ref{table:meta-psnr}. We can observe that the models that are both trained and inferred with our partition methods achieve the best performance, which meets the conclusion in Table~\ref{table:meta-psnr}.

\begin{table}[]\footnotesize
\caption{\footnotesize Mean SSIM and LPIPS with 3 views for learning-to-learn INRs methods. For short, we rewrite
G as the PoG partition method and S as the PoS partition method.
We mark a superscript G/S if we apply the corresponding method
in the training stage and mark a subscript G/S if we apply the cor-
responding method in the fine-tuning stage}
\begin{tabular}{ccc|ccc}
\hline
Setting                              & SSIM $\uparrow$ & LPIPS $\downarrow$ & Setting                              & SSIM $\uparrow$ & LPIPS $\downarrow$ \\ \hline
$\text{SIREN \space}$                & 0.72            & 0.46               & -                                    &                 &                    \\
$\text{SIREN}_{\text{G}}$            & 0.75           & 0.42              & $\text{SIREN}_{\text{S}}$            & 0.76           & 0.36              \\
$\text{SIREN}_{\text{G}}^{\text{G}}$ & \textbf{0.79}  & \textbf{0.33}     & $\text{SIREN}_{\text{S}}^{\text{S}}$ & \textbf{0.79}  & \textbf{0.32}     \\
$\text{SIREN}_{\text{G}}^{\text{S}}$ & 0.78           & 0.36              & $\text{SIREN}_{\text{S}}^{\text{G}}$ & 0.64           & 0.44              \\ \hline
\end{tabular}
\label{table:ssim2}
\end{table}

\end{document}